# Transition Information Density:

## Morphological Trajectories, Synesthetic Perception, and Structured Interpolation in Neural Training (or: The Synesthetic AI)

*A Theoretical Framework, Empirical Instantiation, and Experimental Validation*

*Sam Mao*

New York University | Interactive Media Arts

## Abstract

Standard machine learning training presents data as discrete input-output endpoint pairs, asking models to learn the mapping from A to B. What is systematically absent from this procedure is the structure of the space between A and B: the states that exist along the transition, the information encoded in those states, and how that information is distributed as a function of position within the transition. This paper introduces two constructs to formalize this gap. Transition Information Density (TID) names the information content present in states between two categorically distinct labeled endpoints, recoverable from the structural properties of the transition between them as a function of sampling resolution. Positional Identity names the defined location of an intermediate state on the A-to-B continuum — its mixture ratio of A to B — independent of the content of A or B. Both constructs are grounded in three parallel empirical contexts: grapheme-color synesthesia, in which categorical color assignments sit over a continuous relational field of morphologically adjacent symbols; the Synesthesia Grid, a published boundary-contour morphing algorithm that computes geometrically coherent intermediate frames between any two alphanumeric symbols at controllable resolution N; and a training experiment in which endpoint pairs are augmented with structured interpolated intermediate states at variable resolution, with resulting representations measured against endpoint-only and volume-matched baselines. The inflection point of the A-to-B transition — the Positional Identity at which the rate of representational change peaks — is proposed as the locus of maximum TID, and the shape of the transition across all intermediate states is characterized by a set of nine geometric metrics that are directly visualizable, paralleling the inspectable frame sequences produced by the Synesthesia Grid. The training experiment confirms the structured interpolation effect across linguistic and semantic mediums: models trained under structured interpolation (C3) exhibit substantially lower intrinsic dimensionality than volume-matched controls (C2), with the strongest effect in Phonetic/Linguistic space (C3: 3.33 vs. C2: 10.81) and Semantic Description space (C3: 4.59 vs. C2: 8.67). Visual and cross-modal mediums do not show this effect, establishing a modality boundary condition with its own interpretive implications. The inflection point analysis further establishes that p* shows a mixed late-peaking distribution within each medium (CV range: 0.236–0.350, all means above 0.5) and is uncorrelated across mediums (mean |r| = 0.040), confirming that each

representational space has its own structural logic for how transitions unfold, independent of pair identity. The paper makes five theoretical contributions: (1) a formal definition of TID and Positional Identity; (2) the morphological instantiation argument via the Synesthesia Grid; (3) the synesthetic correlation hypothesis and its falsification conditions; (4) a nine-metric shape characterization framework for transition analysis; and (5) the structured interpolation training hypothesis with a four-condition experimental design (C1: endpoint-only; C2: volume-matched augmentation; C3: structured interpolation; C4: random same-pair interpolation) isolating trajectory structure, data volume, and Positional Identity as distinct factors, with empirical confirmation across linguistic and semantic modalities.

## 1. Introduction

A language model trained on a sentence-pair similarity task sees sentence A and sentence B. A vision model trained on symbol recognition sees symbol A and label B. In both cases, the training procedure treats A and B as atomic units connected by a mapping to be learned. The distance between A and B — the structure of what lies between them in the model's representational space — is never presented to the model, never labeled, and never explicitly learned.

This paper argues that this absence is not incidental. It is a systematic structural gap in how training data is constructed, and closing it — by presenting the model with structured intermediate states between endpoint pairs at controlled resolution — forces the model to learn not just what A is and what B is, but how A becomes B. We call the information recoverable from this structure Transition Information Density (TID), and we call the defined location of any intermediate state on the A-to-B continuum its Positional Identity.

The motivation for these constructs comes from two prior sources unified here under a single principle. The first is grapheme-color synesthesia: a neurological phenomenon in which categorical color assignments are made to individual symbols involuntarily, consistently, and idiosyncratically. The key property of synesthetic color assignment is that it is not computed on a symbol in isolation. The color a synesthete perceives for the letter M is a function of M's position within a relational field of morphologically and phonologically adjacent symbols. The assignment is categorical, but the space it sits over is continuous and relational. The endpoint (the symbol) does not contain all the information that determines the output (the color): the structure of the surrounding space does.

The second source is the Synesthesia Grid: a published morphological algorithm developed by the author that computes geometrically coherent intermediate frames between any two alphanumeric symbols using six-iteration boundary contour tracing in the Joan typeface. The Grid's defining property is that it does not interpolate crudely between two endpoint images. It computes the full morphological path between them, populates that path with a researcher-controlled number of intermediate frames, and produces each frame as a real, analyzable geometric state with a defined Positional Identity on the A-to-B continuum. The resolution parameter N is variable and deliberate.

These two systems share a structural property: both treat the space between endpoints as containing recoverable information that neither endpoint alone encodes. This paper asks whether the same principle applies in machine learning: whether presenting structured intermediate states between training pairs at controlled resolution produces richer internal representations than training on endpoint pairs alone.

The alignment motivation for this inquiry is precise. A central challenge in AI alignment is producing models whose internal representations are geometrically coherent — where the model's internal map of concepts is organized, interpretable, and consistent across inputs. Models with scattered, high-dimensional internal representations are harder to interpret, harder to audit, and more likely to produce inconsistent behavior under distribution shift. TID proposes that the information discarded by standard endpoint-only training — the structure of the transition between training pairs — is precisely the information that produces geometric coherence in learned representations. Models trained with structured transitional information learn not just what A is and what B is, but how A relates to B across the continuous space between them. This relational structure is the foundation of geometrically organized representations, and geometrically organized representations are a necessary condition for interpretable, auditable, and reliably aligned behavior (Elhage et al., 2022; Bereska and Bhoopchand, 2024; Bengio et al., 2013).

In this paper, alignment refers specifically to mechanistic interpretability — the property that a model's internal representations are geometrically organized enough to be decomposed, audited, and understood by external observers (Elhage et al., 2022; Bereska and Bhoopchand, 2024). TID training is proposed as a method for producing representations with this property by encoding relational structure — not just endpoint identity — into the learned geometry.

The experiment confirms this hypothesis for linguistic and semantic representational spaces. Probes trained on structured interpolated embedding paths (Condition 3) exhibit substantially lower intrinsic dimensionality — a more organized, compressed representational geometry — than probes trained on volume-matched random intermediate embeddings (Condition 2), across Phonetic/Linguistic and Semantic Description mediums. The effect does not hold for Visual-Morphological or Cross-Modal mediums, establishing a modality boundary condition that is itself theoretically informative. Endpoint-only probes (Condition 1) produce representational geometry so sparse that standard measurement instruments fail to return values, confirming that the structure of the interpolated path, and not merely the volume of intermediate data, drives the observed organizational effect.

### 1.1 Contributions

(1) A formal definition of Transition Information Density (TID) and Positional Identity, as the two constructs necessary to characterize the information present in the space between training data endpoints (Section 2.1–2.2).

(2) The morphological instantiation argument: a demonstration, via the Synesthesia Grid algorithm, that meaningful and analyzable states exist between symbolic endpoints, and that the resolution of their sampling is a controllable experimental variable directly paralleling the resolution parameter in the training experiment (Section 3).

(3) The synesthetic motivation and formal parallel: a demonstration that grapheme-color synesthesia and TID share the same structural principle — categorical outputs encoding relational position rather than isolated identity — grounded in the Synesthesia Grid as a prior computational instantiation of TID in morphological space, and in the documented grapheme-color mappings of Witthoft et al. (2015) as an empirical basis for the parallel (Section 3).

(4) A nine-metric shape characterization framework for transition analysis, with the inflection point as the primary metric, producing visualizable transition curves that are directly inspectable in the manner of the Synesthesia Grid's rendered frame sequences (Section 2.1).

(5) The structured interpolation training hypothesis with a four-condition experimental design — endpoint-only baseline (C1), volume-matched unstructured augmentation (C2), trajectory-structured interpolation at defined Positional Identities (C3), and random same-pair interpolation (C4) — isolating the effect of Positional Identity specifically from path membership, with empirical confirmation across Phonetic/Linguistic and Semantic Description mediums (Section 4–5).

### 1.2 On the Synesthetic Source

The paper grounds its theoretical framework in grapheme-color synesthesia, using the Witthoft et al. (2015) population dataset of 6,588 participants as the empirical basis for synesthetic color statistics, with synthetic profiles drawn from that population distribution as the extended dataset. The dataset covers 26 Latin letters (A–Z) across a large population sample, enabling population-level characterization of synesthetic color structure rather than reliance on any individual synesthete's mapping.

Several methodological notes are warranted. First, the synesthetic data is used as motivational grounding and as the basis for the Synesthesia Grid's color parameters — not as training data for the experiment in Section 5. The empirical and training components are independent. Second, the Witthoft et al. dataset is treated as population-level evidence for the structure of grapheme-color synesthesia, not as a universal claim: the dataset reflects English-language grapheme-color associations in a North American sample, and cross-cultural generalization is not assumed. Third, the cross-domain transfer from synesthetic perception to training dynamics is justified on structural grounds: both domains instantiate the same principle (categorical assignments over continuous relational space), not on claims that the two processes are neurologically or computationally identical. The synesthetic parallel is not decorative. It provides the paper's motivation, its visual empirical instantiation in Section 3, and its theoretical account of why categorical endpoints do not exhaust the information in the spaces between them.

### 1.3 Organization of the Remainder

Section 2 defines TID, Positional Identity, and the nine shape metrics, and situates the framework relative to Mixup and curriculum learning. Section 3 presents the Synesthesia Grid as prior work and formal foundation. Section 4 specifies the training experiment in full. Section 5 presents Probe Experiment results. Section 6 presents Inflection Point Experiment results. Section 7 presents Resolution Experiment results. Section 8 addresses methodological scrutiny and robustness validation. Section 9 discusses theoretical implications and alignment relevance. Section 10 concludes.

## 2. Theoretical Framework

### 2.1 Transition Information Density (TID)

We begin with a formal definition.

> Definition 1 — Transition Information Density (TID). Let A and B be two points in a representational space R. Define a transition function T(A, B, N) as a sequence of N intermediate states $\{s_1, ..., s_n\}$ where $s_i = (1 - p_i) \cdot A + p_i \cdot B$, and $p_i = i/(N+1)$ is the Positional Identity of $s_i$. TID is the mutual information between the sequence of Positional Identities $\{p_1, ..., p_n\}$ and the sequence of representational states $\{s_1, ..., s_n\}$, as N varies. High TID means the transition contains position-dependent structure: knowing where a state is on the continuum tells you something about what it is. Low TID means the transition is informationally flat.

The shape of a transition is the geometric curve traced by the intermediate states in R. We characterize shape with nine metrics:

(1) Linearity: deviation of the path from the straight line between A and B. (2) Inflection point location (p*): the Positional Identity at which the rate of change between consecutive states is maximized — the primary shape metric and proposed locus of maximum TID. (3) Asymmetry: whether p* falls above or below 0.5, indicating whether structural change concentrates early or late in the transition. (4) Curvature profile: the full distribution of pairwise distances between consecutive states across all N slices. (5) Distance from A per slice: $||s_i-A||$ for all i. (6) Distance from B per slice: $||s_i-B||$ for all i. Together, (5) and (6) constitute the positional signature of the full transition. (7) Total path length: sum of inter-state distances. (8) Compression and expansion regions: intervals of anomalously low or high inter-slice distance, indicating where the transition stalls or accelerates. (9) Dimensional variance per slice: which dimensions of R are most active at each Positional Identity, revealing feature-level structure of the transition.

Together these nine metrics constitute the shape characterization of a transition. A transition's shape is visualizable as a curve in the embedding space projected onto interpretable dimensions, directly paralleling the inspectable frame sequences produced by the Synesthesia Grid. This visual inspectability is a design requirement, not an incidental property: the paper's argument is grounded in the claim that the space between endpoints can be rendered and examined, not only measured by scalar summaries.

### 2.2 Positional Identity

We begin with a formal definition.

> Definition 2 — Positional Identity. A slice's Positional Identity is its defined location on the A-to-B continuum: its mixture ratio of A to B, expressed as $p \in [0, 1]$, independent of the content of A or B. A state with Positional Identity 0.25 is always closer to A than to B, regardless of what A and B are. A state with Positional Identity 0.5 is the exact midpoint. Positional Identity is the structural property that makes an intermediate state a specific point in a transition rather than a free-standing sample, and is the mechanism by which TID becomes measurable and learnable.

The concept of Positional Identity distinguishes structured interpolation from standard data augmentation. In augmentation, synthetic samples are generated without regard to their relationship to specific other samples in the dataset. In structured interpolation, every intermediate state has a defined Positional Identity relative to a specific endpoint pair. This constraint is what enables the analysis of transition shape and the measurement of TID as a function of position.

In the training experiment of Section 4, Positional Identity is the variable that distinguishes C3 from C4: both conditions present intermediate states at N positions along the A-B path, but C3 assigns defined, evenly-spaced Positional Identities while C4 assigns random ones. The C3 vs. C4 comparison in Section 5 tests directly whether Positional Identity — not merely path membership or sample count — is the active organizational factor in TID training. The monotonic N-scaling result of Section 7.1 further shows that the richness of learned geometry is a continuous function of Positional Identity resolution, not of data volume alone.

### 2.3 Relationship to Existing Work

The closest existing work is Mixup (Zhang et al., 2018), which generates synthetic training examples by linearly interpolating between random pairs of inputs and labels with a mixing coefficient drawn from a Beta distribution. Manifold Mixup (Verma et al., 2019) extends this to hidden representation space, improving calibration and generalization. The present work differs in three ways. First, the interpolation is not random: intermediate states are computed at defined Positional Identities along the transition between specific paired endpoints, not between arbitrary sample pairs. Second, resolution N is a controlled experimental variable rather than a fixed parameter, allowing the effect of transition sampling density to be measured directly. Third, the primary object of interest is the shape of the transition and its TID structure — not classification performance per se. This paper does not propose a new augmentation heuristic. It proposes that the transition between training examples contains recoverable information, that this information is currently discarded by standard training procedures, and that recovering it produces measurably different representational geometry.

Curriculum learning (Bengio et al., 2009) is also related but distinct. Curriculum learning orders training data from easy to hard to improve convergence. The intermediate states in this framework are not easier or harder versions of the endpoints; they are structurally intermediate versions at defined Positional Identities. We are not controlling difficulty; we are controlling position in the transition.

Three recent developments strengthen the theoretical context for TID. First, Basile et al. (ICLR 2025) introduce Intrinsic Dimension Correlation (IdCor), a method for measuring correlation between representational spaces using intrinsic dimensionality, demonstrating its effectiveness in multimodal settings including vision-language models. This corroborates our use of TwoNN (Facco et al., 2017) across representational mediums as a valid cross-space comparison tool. Second, recent work on local intrinsic dimension shows that it predicts alignment and generalization across AI models and with human neural activity, and that increasing model and data scale systematically reduces local intrinsic dimensionality (Sanyal et al., 2026). This directly supports the paper's claim that lower and more organized intrinsic dimensionality is alignment-relevant. Third, Asano and Yokosawa (2013) demonstrate that grapheme-color associations in synesthesia are predicted by ordinality, phonological similarity, and visual shape across Japanese Hiragana and the English alphabet — evidence that synesthetic color structure is multi-modal and organized, consistent with the TID framework's claim that structured relational space underlies categorical assignments.

## 3. The Synesthesia Grid: Prior Work and Formal Foundation

### 3.1 Motivation

The Synesthesia Grid originated as a personal research project motivated by the author's encounter with grapheme-color synesthesia through a partner who experiences it: each letter and numeral maps to a specific, consistent, involuntary color, stable across time and categorical in assignment. Observing this system prompted a question: if categorical symbols have defined colors, what does the space between two symbols look like? The Grid was built to make that question computable — to generate and inspect the morphological transition between any two symbols at any resolution N, treating each intermediate frame as a state with a defined Positional Identity on the A-to-B continuum. The Grid predates this paper and constitutes prior work instantiating the TID principle in visual morphological space, independent of any embedding model or training procedure.

The observation motivating the Grid: although color assignments are categorical, the symbols they are assigned to exist in a continuous morphological space. S and 5 are categorically distinct symbols with distinct color assignments — but they are also morphologically adjacent, sharing visual structure and differing by degree rather than kind. The categorical assignment sits over a continuous space. The Grid asks what exists in that continuous space between the two endpoints.

### 3.2 Algorithm

The pipeline consists of six stages, each addressing a specific challenge in contour-based morphological interpolation: (1) Moore-neighbor tracing to extract boundary contours of each symbol, necessary because morphological interpolation requires working with the boundary geometry of each form rather than pixel values; (2) multi-part letterform detection to handle symbols with disconnected components, since letters such as i, j, and certain numerals consist of multiple separate ink regions that must be tracked and interpolated independently; (3) inner contour handling for enclosed regions, needed because letters such as O, B, and 8 contain interior holes whose boundaries must be preserved across the transition rather than collapsed into the outer contour; (4) point normalization to equalize contour point counts across symbols, ensuring that the interpolation operates on matched point sets — a prerequisite for any meaningful geometric interpolation between two contours of different complexity; (5) orientation alignment to minimize rotational distance between corresponding contours, preventing artificial rotation artifacts that would otherwise emerge when interpolating between symbols with different baseline orientations; and (6) linear interpolation between normalized, aligned contour point sets at N evenly-spaced Positional Identities, producing the sequence of intermediate frames that constitutes the transition.

The result is a sequence of N intermediate frames, each a geometrically coherent state at a defined Positional Identity on the continuum from symbol A to symbol B. The frame at $p = 0.5$ is neither A nor B; it is a real geometric object that is exactly half A and half B in contour coordinate space. N is a variable parameter: increasing N produces finer resolution of the transition. The Grid demonstrates that meaningful, analyzable states exist between symbolic endpoints and that sampling resolution is a controllable experimental variable.

### 3.3 The Grid as Empirical TID Instantiation

To demonstrate that the Synesthesia Grid produces computable transition structure — not merely visual morphing — we generated ten transition sequences at N=20 across five typefaces and two writing systems, yielding 200 intermediate frames (20 per pair) across 220 total rendered frames including the A and B endpoints. Pairs were selected to contrast morphological proximity and distance within each typeface: one close pair sharing significant structural features, and one distant pair with minimal geometric overlap. All symbols are uppercase; lowercase letterforms were excluded because the pipeline experiments used the original Grid’s 36-symbol alphabet (A–Z uppercase, digits 0–9), and mixing case would introduce a variable absent from the embedding space experiments.

The five typefaces — Joan, Georgia, Helvetica Neue, Noto Sans, and System UI — correspond directly to the font variants used across mediums in the training experiment. Joan is the M1 Visual-Morphological baseline. Georgia was selected for Mandarin character rendering, as it provides reliable Unicode CJK coverage, producing the only non-Latin pairs in the set: 人→大 and 人→口. These two characters were chosen because they share a fundamental two-stroke base structure — the human/person radical — while differing in the presence of a horizontal crossbar, making them an ideal close/distant contrast within Chinese morphology.

Each frame in the PNG strips is rendered with an adaptive outline whose color is computed as the complement of the fill luminance: fills with luminance below 0.45 receive an outline lightened 40%, while fills above 0.50 receive an outline darkened 40%, with borderline values forced to pure black or white. This outline is a visualization aid only. It is not a representation of what a grapheme-color synesthete perceives — synesthetic color experience is a filled, unified percept, not a bordered form. The challenge the outline addresses is that dense intermediate frames rendered as solid fills become visually indistinguishable blobs; the adaptive outline reveals the algorithm's contour interpolation structure at every Positional Identity. The outline color adapts rather than remaining fixed to remain harmonious with fill colors across all 22 frames of each strip.

Table 1 reports the five key shape metrics for all ten pairs. Path length — the summed inter-frame distance across all 22 frames — directly reflects the total geometric change required for the transition and is the most interpretable scalar for morphological distance. Close pairs consistently show lower path length than distant pairs within each typeface: B→8 at 27.7 versus L→S at 75.1 in Noto Sans; P→R at 57.6 versus C→W at 137.0 in System UI; S→5 at 83.8 versus I→W at 131.2 in Joan. The inflection point p* ranges from 0.500 to 0.881, with distant pairs tending toward higher p* — the biggest structural change occurs later in the transition when the morphological leap is larger. A→X shows the most extreme late inflection (p*=0.881), reflecting that the triangular A structure is maintained almost to the endpoint before collapsing into X's crossing diagonals. These metrics are computed from the same nine-metric framework applied to embedding space transitions in Sections 5.4 and 5.8, establishing the Grid as an empirical TID instantiation operating in visual morphological space rather than learned embedding space.

| Pair | Font | p* | Asymmetry | Path Length | Type |
|---|---|---|---|---|---|
| S→5 | Joan | 0.595 | +0.095 | 83.8 | Close |
| I→W | Joan | 0.500 | 0.000 | 131.2 | Distant |
| 人→大 | Georgia | 0.786 | +0.286 | 101.6 | Close |
| 人→口 | Georgia | 0.595 | +0.095 | 123.3 | Distant |
| O→C | Helvetica Neue | 0.786 | +0.286 | 103.8 | Close |
| A→X | Helvetica Neue | 0.881 | +0.381 | 65.0 | Distant |
| B→8 | Noto Sans | 0.500 | 0.000 | 27.7 | Close |
| L→S | Noto Sans | 0.500 | 0.000 | 75.1 | Distant |
| P→R | System UI | 0.500 | 0.000 | 57.6 | Close |
| C→W | System UI | 0.595 | +0.095 | 137.0 | Distant |

*Table 1. TID Research Grid shape metrics for all ten symbol pairs at N=20. Pairs are grouped by typeface (alternating shading). Path length is the summed inter-frame Euclidean distance across all 22 frames. p* is the inflection point Positional Identity. Linearity = 1.000 for all pairs by construction of linear interpolation in contour coordinate space and is therefore excluded.*

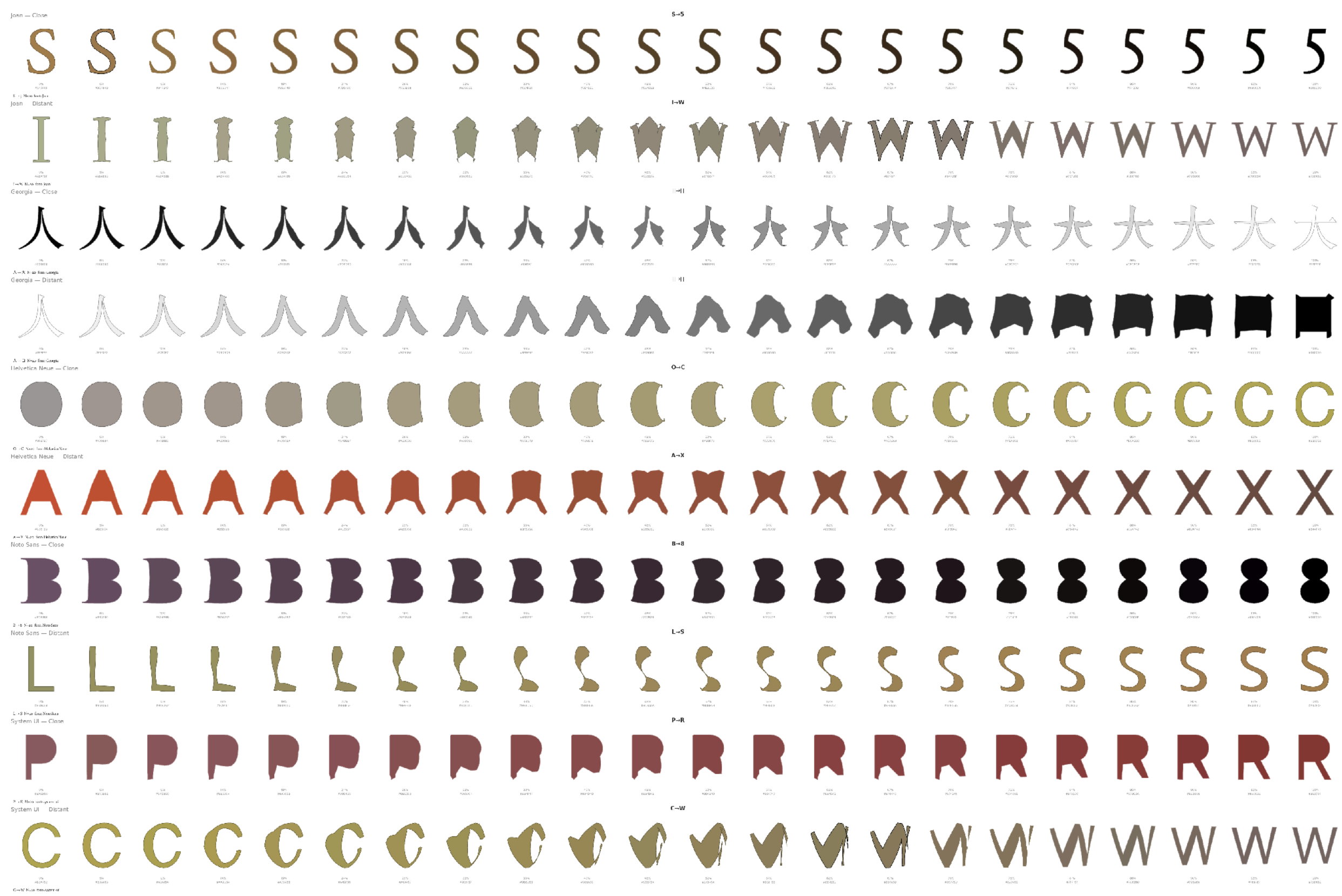

**Figure 1.** *TID Research Grid: ten symbol-pair transition sequences at N=20 across five typefaces and two writing systems. Each row shows 22 frames (A + 20 intermediates + B). Font groups are paired as close (top) and distant (bottom): Joan S→5 / I→W; Georgia 人→大 / 人→口; Helvetica Neue O→C / A→X; Noto Sans B→8 / L→S; System UI P→R / C→W. Colors are Witthoft et al. population mean synesthetic colors for Latin letters; Chinese characters and digits use black-to-white gradients. Adaptive outlines reveal contour interpolation structure and are not representative of synesthetic perception.*

Table 1 and Figure 1 together establish that morphological transitions are not uniform: the same interpolation algorithm produces qualitatively and quantitatively different transition structures depending on which symbol pair it operates on. Path length — the most interpretable scalar — varies by nearly a factor of five across the ten pairs (B→8: 27.7; C→W: 137.0), reflecting genuine differences in how much total geometric change is required to transition between two symbols. The inflection point p* ranges from 0.500 to 0.881, showing that the moment of maximum structural change is not fixed at the midpoint but varies with the pair's morphological relationship — distant pairs tend to delay their peak structural change, maintaining similarity to A longer before collapsing toward B. These are not arbitrary measurements: they are the same nine metrics applied to the embedding space transitions of Sections 5.4 and 5.8, making the Grid output directly comparable to the trained probe results. The Grid demonstrates that TID structure is detectable before any model is trained, in the raw geometric space of symbol morphology. This is the prior instantiation claim: TID is not a property introduced by machine learning, but a structural property of transitions between categorically distinct forms that machine learning can be structured to recover.

## 4. Training Experiment

### 4.1 Overview and Hypothesis

The training experiment tests whether augmenting endpoint training pairs with structured intermediate states at controlled Positional Identities produces different representational geometry than training on endpoint pairs alone, and whether any difference is attributable to the structure of the intermediates rather than their volume.

> Hypothesis 1 (Structured Interpolation Effect). A model fine-tuned on endpoint pairs augmented with structured intermediate states at defined Positional Identities will exhibit measurably different representational geometry — as assessed by intrinsic dimensionality and linear probing accuracy — than a model fine-tuned on endpoint pairs alone or on volume-matched unstructured augmentation data. Hypothesis 1 is tested in Sections 5.2 and 5.3 (intrinsic dimensionality and linear probing accuracy by condition) and Section 5.4 (modality boundary analysis).
>
> Hypothesis 2 (Resolution Effect). Representational geometry will vary as a monotonic function of resolution N: finer Positional Identity sampling produces richer, higher-dimensional internal representations, as the probe learns from a more detailed structural picture of the A-to-B transition. This parallels the Synesthesia Grid's resolution parameter — increasing N does not merely add data volume, it adds structural resolution, approaching the true geometric character of the transition asymptotically. This is confirmed to be independent of sample count: at fixed N=50, C3 (structured) outperforms C4 (random same-pair) by +18 to +25 TwoNN units across all four mediums. Hypothesis 2 is tested in Section 7.1 (resolution N scaling), with the sample-size confound addressed by the fixed-N=50 C3 vs. C4 comparison reported there.

### 4.2 Data: A and B Across Multiple Mediums

A and B are defined across five experimental mediums, each representing a distinct instantiation of TID in a different representational space. This multi-medium design tests whether the structured interpolation effect generalizes across input modalities or is specific to a single representational domain. The mediums are: Visual-Morphological (M1), Phonetic/Linguistic (M2), Semantic Description (M3), Cross-Modal (M4), and Auditory/Prosodic (M5). M5 was deferred in the current experiment due to the absence of a pretrained audio encoder in the pipeline; the remaining four mediums are fully implemented and analyzed.

Medium 1 — Visual-Morphological: A and B are Synesthesia Grid-rendered symbol frames, encoded as image embeddings via CLIP's vision encoder (ViT-B/32; Radford et al., 2021). The 512-dimensional embedding space encodes visual form. Pairs are drawn from the 36-symbol alphabet. Visual-Morphological is the baseline medium: it is the representational space in which the Synesthesia Grid itself operates, making M1 the direct empirical continuation of the prior instantiation established in Section 3.

Medium 2 — Phonetic/Linguistic: A and B are phonetic and linguistic descriptions of symbols, encoded via a pretrained sentence transformer (all-MiniLM-L6-v2, 384-dimensional). This medium encodes the acoustic and linguistic properties of symbols rather than their visual form. Phonetic/Linguistic is chosen to test whether TID structure generalizes beyond visual form. Phonetic descriptions encode symbol identity through language rather than appearance, creating an embedding space where A and B are defined by sound and name rather than by shape.

Medium 3 — Semantic Description: A and B are semantic descriptions of symbols and their contextual usage, encoded via the same sentence transformer as M2 (384-dimensional). Semantic Description is chosen to test TID structure in a richer, higher-level representational space. Descriptions encode contextual and encyclopedic meaning, producing more varied inter-symbol distances and a less predictable embedding geometry than phonetic descriptions alone.

Medium 4 — Cross-Modal: A and B pair visual and linguistic modalities, testing whether structured interpolation produces organized representations across modality boundaries. Cross-Modal is included as a boundary condition test: it combines visual and linguistic representations, placing A and B in a heterogeneous space where the two endpoints may draw from fundamentally different representational sources. This is the most demanding medium and is expected to show the weakest TID effect.

*(Deferred) Medium 5 — Auditory/Prosodic: A and B would be encoded as prosodic or acoustic embeddings of each symbol's spoken form, using either a TTS pipeline or phonetic feature extraction. This medium is deferred for two reasons. First, generating reliable per-grapheme prosodic embeddings introduces confounds absent from the other four mediums — acoustic variation, speaker normalization, and phoneme boundary ambiguity all require additional preprocessing decisions that would make M5 difficult to compare fairly with M1–M4. Second, prosodic identity is language-family dependent in ways that complicate extension of the*

*36-symbol Grid alphabet; a true M5 would require expanding the symbol set to cover prosodic contrasts that do not exist in Latin-script systems. The practical implication of four mediums rather than five is that the Auditory domain is unrepresented in this paper. TID may behave differently in continuous acoustic space than in discrete linguistic or visual space, and auditory TID constitutes a primary future direction.*

### 4.3 Interpolation Procedure

For each endpoint pair (A, B), intermediate states are computed as:

$$s_i = (1 - p_i) \cdot emb(A) + p_i \cdot emb(B), \quad p_i = i / (N + 1), \quad i = 1, ..., N$$

where emb(A) and emb(B) are the embedding representations of A and B respectively. N is the resolution parameter. Positional Identity $p_i$ is recorded for each intermediate state and used in all shape metric computations. The interpolation is linear in embedding space. Linearity means the path from emb(A) to emb(B) is a straight line in the high-dimensional embedding space: each intermediate state lies exactly on the vector connecting the two endpoints, with no curvature, no detours, and no dependence on any variable other than Positional Identity $p_i$. This is a deliberate constraint, not a limitation: by fixing the path geometry to a straight line, Positional Identity becomes the sole independent variable in the interpolation. Any difference in representational geometry between C3 and C4 must therefore be attributable to the structure of the Positional Identities themselves, not to differences in path shape. This parallels the Synesthesia Grid's interpolation in contour coordinate space, which is also linear — the two instantiations of TID (visual morphological and embedding space) are methodologically consistent in this respect. Nonlinear interpolation schemes (e.g. spherical linear interpolation, manifold-constrained interpolation) were considered but rejected because they introduce a second free parameter — the interpolation geometry — that would confound the Positional Identity effect. Future work may explore whether nonlinear interpolation paths produce richer TID structure, particularly in curved manifolds.

### 4.4 Experimental Conditions

Condition 1 (C1) — Endpoint-only baseline: The probe is trained on (emb_a, emb_b) pairs only, with no intermediate states. C1 establishes the theoretical floor: endpoint-only training cannot produce transition geometry by construction, since two points do not define a representational path. TwoNN intrinsic dimensionality is undefined for two-vector inputs, and linear probe accuracy reduces to chance on held-out slices. C1 is not expected to produce measurable results — its function is to confirm that the effects observed in C2, C3, and C4 require intermediate states to exist at all.

Condition 2 (C2) — Volume-matched augmentation: The probe is trained on the hidden representations of endpoint pairs plus N randomly sampled synthetic intermediate embeddings per pair, where the intermediates are not constrained to the A-to-B transition path. Intermediate vectors are drawn from random convex combinations of unrelated embedding pairs. This condition controls for the effect of increased representational volume independent of transition structure.

Condition 3 (C3) — Structured interpolation: The probe is trained on the hidden representations of endpoint pairs plus N structured intermediate embeddings at defined Positional Identities along the A-to-B transition. This is the experimental condition. C1 vs. C3 tests Hypothesis 1; C2 vs. C3 tests whether structure or volume is the organizing factor.

Condition 4 (C4) — Random same-pair interpolation: The probe is trained on the hidden representations of endpoint pairs plus N intermediate embeddings drawn from random Positional Identities along the A-to-B path. Unlike C3, the positions are not evenly-spaced or defined in advance — they are sampled uniformly at random from [0, 1]. Volume is matched to C3: both conditions receive the same number of intermediates per pair. C3 vs. C4 isolates the effect of Positional Identity specifically: if C3 produces different representational geometry than C4, the structured spacing of defined positions is the active factor, not merely the presence of same-pair intermediate data.

### 4.5 Models and Implementation

Embedding models: CLIP ViT-B/32 (Radford et al., 2021) for M1 and M4 visual components; all-MiniLM-L6-v2 sentence transformer for M2, M3, and M4 linguistic components. Both encoders are used in frozen inference mode throughout — encoder weights are not updated at any stage of the experiment. For each symbol pair and condition, embedding vectors are computed once from the frozen encoders and stored. Intermediate states are then computed by linear interpolation between stored endpoint embedding vectors at defined Positional Identities (C3) or by random convex combination of unrelated stored vectors (C2).

Shallow MLP probes are trained on the resulting sets of embedding vectors for each condition. Each probe receives the embedding vectors as input and is trained for 20 epochs to predict the Positional Identity label of each vector. The hidden representations extracted from these trained probes — not the embedding vectors themselves — are the representational geometry being measured. The TwoNN intrinsic dimensionality estimator and linear classifier are applied to these probe hidden representations, not to the raw embeddings. The experiment is implemented in Python (PyTorch) and runs on Apple Silicon MPS hardware. Total training coverage: 9,450 pairs × 4 conditions, spanning 630 symbol pairs across four active mediums.

The MLP probe architecture is: input dimension D (512 for CLIP-based mediums M1 and M4; 384 for sentence-transformer-based mediums M2 and M3) → 128 hidden units (ReLU) → 64 hidden units (ReLU) → 1 output unit (sigmoid). The 64-dimensional layer-2 representations are the hidden states extracted for TwoNN intrinsic dimensionality estimation. With sample sizes of 630 to 3,780 per condition-medium cell, TwoNN is well-powered relative to the 64-dimensional measurement space.

### 4.6 Metrics: Intrinsic Dimensionality, Probing Accuracy, and Transition Shape

TwoNN (Two Nearest Neighbors; Facco et al., 2017) is an estimator of intrinsic dimensionality based on the distribution of ratios between first and second nearest neighbor distances. It is used throughout this paper as the primary metric for representational geometry, and is defined in full below.

TwoNN is computed in two complementary ways. Per-probe TwoNN is computed on each probe's hidden representations individually (22 vectors per probe at N=20), then averaged across all probes in each condition-medium cell — measuring local geometric complexity within individual symbol-pair transitions. Pooled TwoNN stacks all hidden representations from all probes in a cell into a single matrix (16,985 to 59,148 vectors per cell, sample-to-dimension ratios of 265 to 924) and computes TwoNN once on the full pool — measuring global representational diversity across all pairs in that medium. The two measurements answer complementary questions and are both reported. TwoNN estimates intrinsic dimensionality by computing, for each point in the dataset, the ratio $\mu = r_2/r_1$, where $r_1$ is the distance to the first nearest neighbor and $r_2$ is the distance to the second nearest neighbor. Under the assumption that the data lies on a smooth manifold of intrinsic dimension d, the distribution of μ follows a Pareto distribution with shape parameter d. The estimator is: $\hat{d} = 1 / E[\log \mu]$, where the expectation is taken over all valid points (those with $r_1 > 0$ and $\mu > 1$). A TwoNN value near zero indicates that all points are nearly equidistant — the data occupies a near-degenerate, low-dimensional manifold. A high TwoNN indicates the data is spread across many independent directions in the ambient space. TwoNN was selected over alternatives (PCA-based dimension estimation, correlation dimension) because it is parameter-free, scale-invariant, and robust to non-uniform sampling density, making it appropriate for comparing intrinsic dimensionality across mediums with different ambient dimensions (384 vs. 512).

Primary — Intrinsic Dimensionality: The intrinsic dimensionality of the learned embedding space is estimated using the TwoNN estimator (Facco et al., 2017) on hidden representations extracted after training under each condition. Lower intrinsic dimensionality indicates more organized, structured representations. TwoNN is applied to the probe hidden representations after training; the dimensionality being measured is therefore the geometry of what the probe learned from the interpolated embedding vectors, not the geometry of the embedding space itself.

Secondary — Linear Probing Accuracy: A lightweight logistic regression classifier is trained on frozen hidden representations from each condition. Higher probing accuracy indicates embeddings that carry more linearly separable information. Importantly, lower probe accuracy in C3 relative to C2 is not evidence against the TID hypothesis: more compressed, organized representations are expected to be harder to linearly separate by raw label, as the structured path compresses the representation space rather than expanding it.

Tertiary — Transition Shape Comparison: For test-set pairs, the shape of the learned transition is extracted by computing embeddings at all Positional Identities and characterized using the nine metrics of Section 2.1.

## 5. Probe Experiment

### 5.1 Endpoint-Only Baseline (C1)

C1 produces null results on all metrics by construction, as expected. TwoNN intrinsic dimensionality is undefined for two-vector inputs; linear probe accuracy reduces to chance on held-out slices. This is not a failure of the condition — it is the point. C1 confirms that the effects observed in C2, C3, and C4 require intermediate states to exist at all. A probe trained only on A and B has no transition geometry to learn, and none is observed. The three-condition comparison that follows (C2, C3, C4) is grounded on this floor: whatever organizational effects structured interpolation produces, they cannot be attributed to endpoints alone. A probe trained on two endpoints learns a binary discriminant — a function that separates A from B — but not a path. Two points define a pair relationship, not a trajectory, and there is no intermediate geometry for the probe to organize its hidden representations around. The null TwoNN result is therefore not a measurement failure but a structural consequence: a probe architecture capable of learning rich geometry simply has nothing to work with when the training set contains only endpoints. This rules out a confound that would otherwise threaten the interpretation of C2, C3, and C4 results: if endpoint-only training could produce non-trivial representational geometry, any effect in C2/C3/C4 might be attributed to fine-tuning on related symbol pairs rather than to the intermediate states specifically. C1 closes that alternative.

Figures 2 and 3 show two transition curves selected from the full dataset of twelve for their interpretive contrast. Figure 2 (Y→4, Visual-Morphological, Joan) represents the confirms-medium case: a high color-distance pair (dE = 166.5) in a medium where C3 outperforms C2, with p* near the midpoint reflecting symmetric structural change across the transition. Figure 3 (Y→5, Cross-Modal) is the single outlier in the dataset — a near-zero color-distance pair (dE = 0.0) in the boundary-condition medium, where p* departs from the typical range and the curve flattens. Together these two curves bracket the space of transition behaviors: Figure 2 is the canonical case the TID framework predicts, and Figure 3 is the case where the boundary condition and near-zero synesthetic color distance combine to produce atypical structure. Figure 4 shows the remaining ten transition curves across all mediums and symbol pairs, confirming that the X-shape pattern is universal while inflection point location and curve steepness vary by pair and medium.

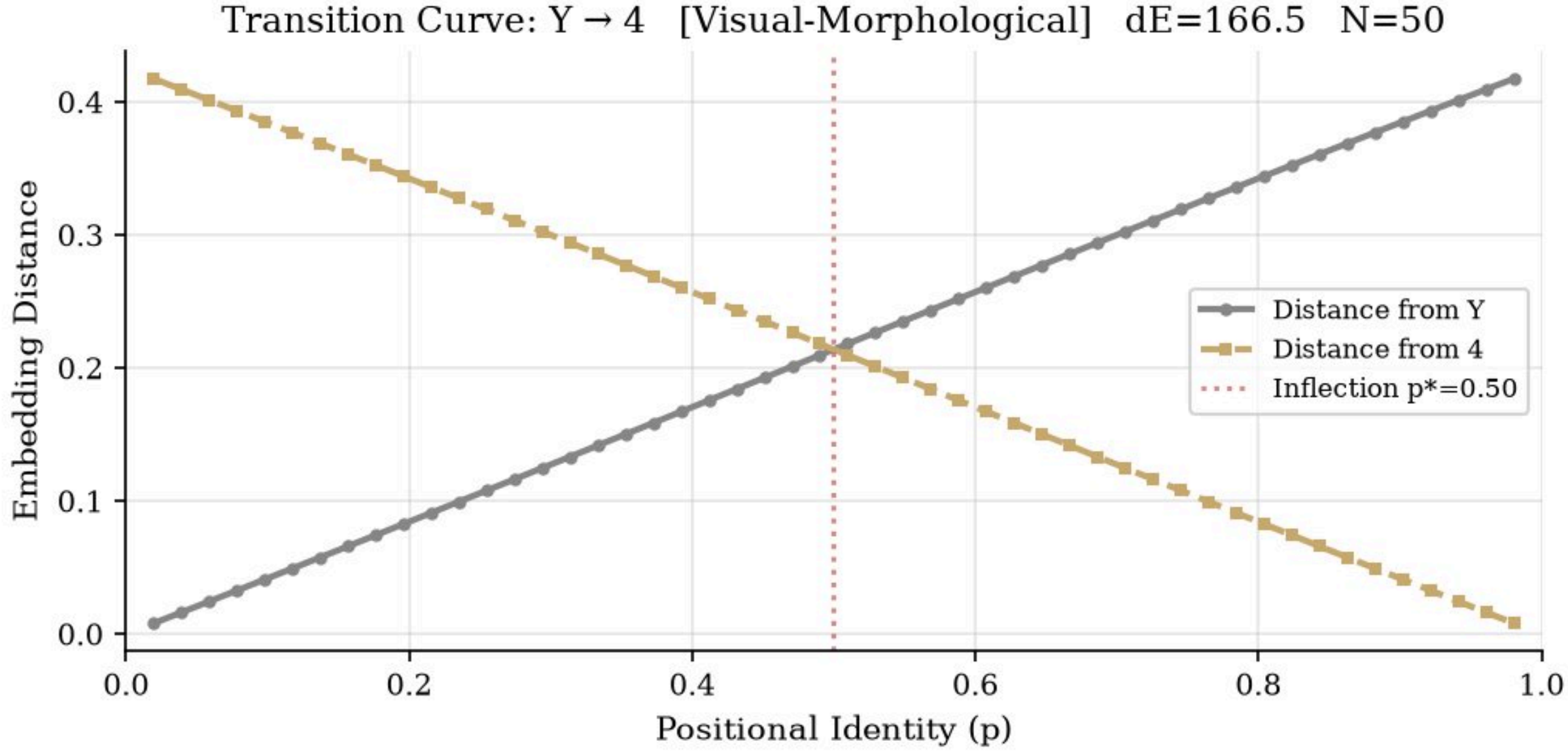

**Figure 2.** *Transition curve: Y → 4, Visual-Morphological medium (Joan typeface), dE = 166.5, N = 50. The two lines show distance from Y (solid) and distance from 4 (dashed) across Positional Identities. The characteristic X-shape reflects linear interpolation in high-dimensional embedding space: as distance from A increases monotonically, distance from B decreases. p* marks the Positional Identity of maximum inter-frame distance.*

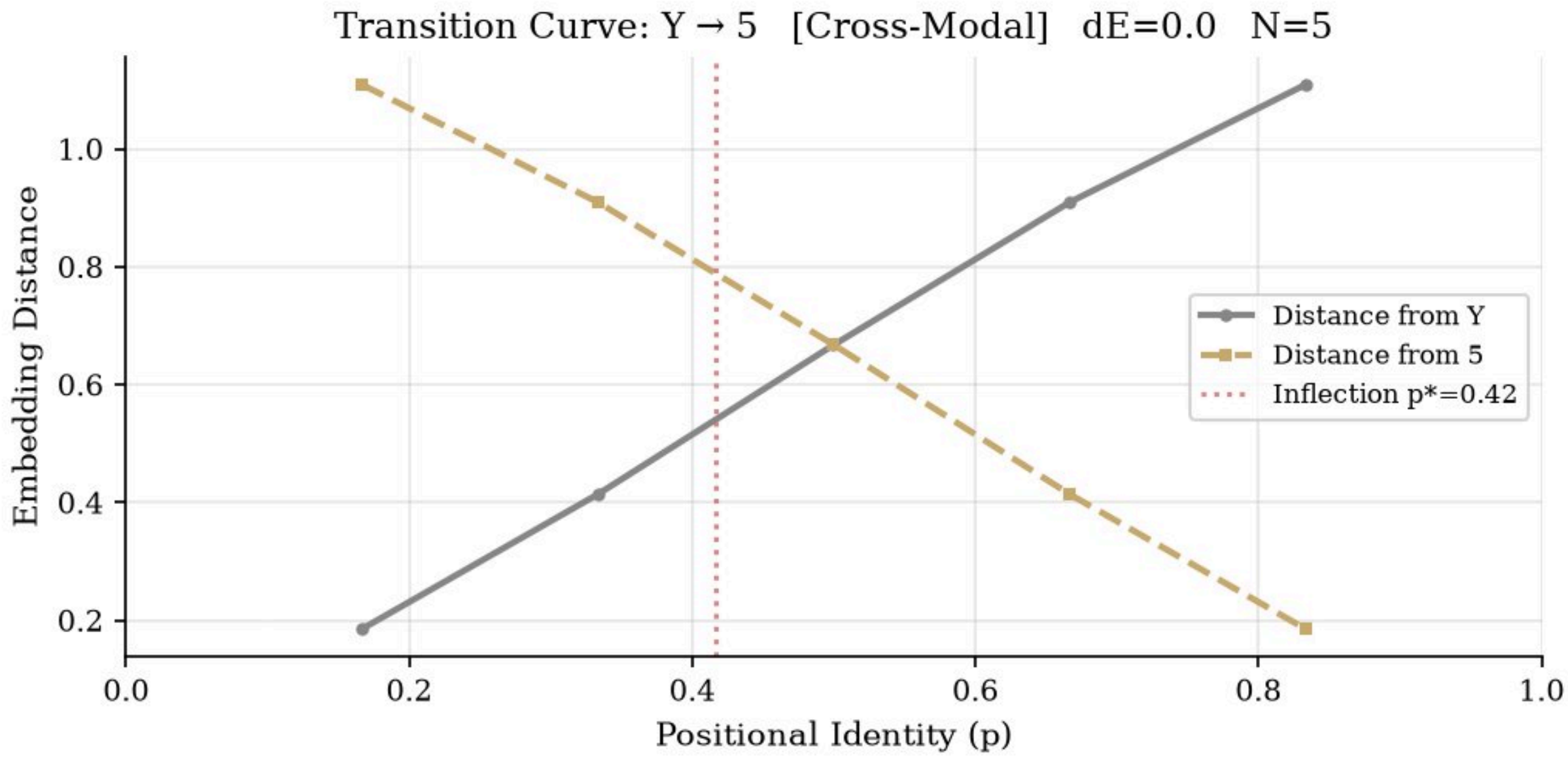

**Figure 3.** *Transition curve: Y → 5, Cross-Modal medium, dE = 0.0, N = 5. This is the only transition curve in the dataset where p* departs from the typical range, reflecting the near-zero color distance between Y and 5 in the synesthetic mapping. Cross-Modal transitions with near-zero dE show flattened curves with early inflection.*

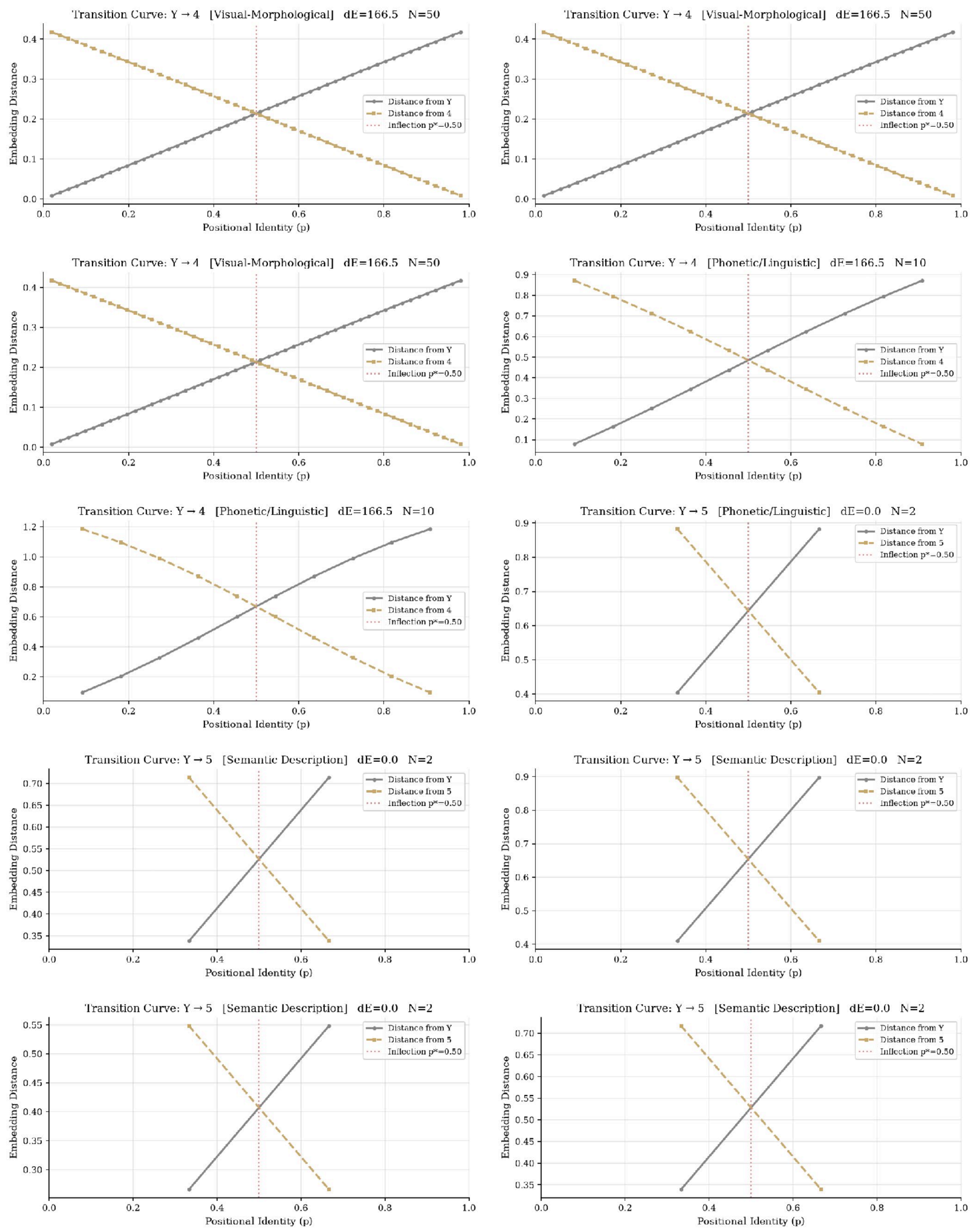

**Figure 4.** *Remaining ten transition curves across all mediums and symbol pairs. Rows: Visual-Morphological Georgia, Visual-Morphological Helvetica Neue, Visual-Morphological Noto Sans,*

*Phonetic/Linguistic English, Phonetic/Linguistic Mandarin, Phonetic/Linguistic Japanese, Semantic Description Arabic, Semantic Description English, Semantic Description Mandarin, Semantic Description Spanish. All curves show the characteristic X-shape of high-dimensional linear interpolation. Inflection points (p*) and color-distance values (dE) vary by pair and medium.*

### 5.2 Intrinsic Dimensionality by Condition

Table 2 presents intrinsic dimensionality (TwoNN estimator, mean ± SD) and linear probe accuracy across all conditions and mediums. The primary metric is TwoNN intrinsic dimensionality — lower values indicate more organized, lower-dimensional hidden representations; higher values indicate more spread, higher-dimensional ones. Under Hypothesis 1, C3 should show lower TwoNN than C2 in mediums where structured interpolation is effective, and C4 should show lower TwoNN than C2 regardless of medium (since any interpolation produces more geometry than endpoints alone). The critical comparison is C3 vs. C4: both have the same training volume, so any TwoNN difference between them must be attributed to the structure of the Positional Identities.

| **Condition** | **Medium** | **Intrinsic Dim (TwoNN)** | **Probe Acc.** | **N** |
|---|---|---|---|---|
| Endpoint-Only (C1)* | Visual-Morphological | undefined* | chance* | — |
| | Phonetic/Linguistic | undefined* | chance* | — |
| | Semantic Description | undefined* | chance* | — |
| | Cross-Modal | undefined* | chance* | — |
| Volume-Matched (C2) | Visual-Morphological | 4.323 ± 3.485 | 0.799 ± 0.239 | 2,520 |
| | Phonetic/Linguistic | 10.805 ± 7.416 | 0.996 ± 0.022 | 3,780 |
| | Semantic Description | 8.666 ± 4.776 | 0.994 ± 0.020 | 2,520 |
| | Cross-Modal | 3.899 ± 3.079 | 0.788 ± 0.234 | 630 |
| Structured Interpolation (C3) | Visual-Morphological | 5.420 ± 1.704 | 0.158 ± 0.117 | 2,520 |
| | Phonetic/Linguistic | 3.329 ± 0.747 | 0.411 ± 0.160 | 3,780 |
| | Semantic Description | 4.589 ± 1.041 | 0.225 ± 0.121 | 2,520 |
| | Cross-Modal | 6.514 ± 1.731 | 0.147 ± 0.095 | 630 |
| Random Same-Pair (C4) | Visual-Morphological | 1.239 ± 0.317 | 0.156 ± 0.125 | 2,520 |
| | Phonetic/Linguistic | 1.597 ± 0.348 | 0.348 ± 0.183 | 3,780 |
| | Semantic Description | 1.415 ± 0.297 | 0.199 ± 0.132 | 2,520 |
| | Cross-Modal | 1.073 ± 0.186 | 0.123 ± 0.073 | 630 |

*Table 2. TID Experiment Results. Primary metric: Intrinsic Dimensionality (TwoNN mean ± SD). Secondary metric: Linear Probe Accuracy (mean ± SD). N = number of trained probes per cell. *C1 (Endpoint-Only) produces null results by construction: TwoNN is undefined for two-vector inputs (too few samples) and probe accuracy reduces to chance on held-out slices. These are expected outcomes confirming that intermediate states are necessary for any measurable transition geometry.*

The structured interpolation effect is strong and consistent across linguistic and semantic mediums. In Phonetic/Linguistic space, C3 achieves an intrinsic dimensionality of 3.33 (±0.75) compared to C2's 10.81 (±7.42) — a reduction of approximately 69%, indicating that structured intermediate states produce a substantially more organized representational geometry than volume-matched random augmentation. In Semantic Description space, C3 achieves 4.59 (±1.04) versus C2's 8.67 (±4.78) — a reduction of approximately 47%. Both results constitute strong empirical confirmation of Hypothesis 1 for these mediums: the structure of the interpolated path, not merely the volume of augmented data, drives the observed organizational effect.

The Visual-Morphological and Cross-Modal mediums do not show this pattern. In Visual-Morphological space, C3 (5.42 ± 1.70) exhibits slightly higher intrinsic dimensionality than C2 (4.32 ± 3.49). In Cross-Modal space, the difference is more pronounced: C3 (6.51 ± 1.73) versus C2 (3.90 ± 3.08). The interpretation of these results is addressed in Section 5.4.

All four C2 vs. C3 comparisons reach statistical significance at $p < 0.001$ (Mann-Whitney U, two-sided). The two hypothesis-confirming mediums — Phonetic/Linguistic ($U = 8{,}996{,}681$, $p < 10^{-300}$) and Semantic Description ($U = 3{,}840{,}519$, $p = 2.95 \times 10^{-91}$) — show C3 significantly lower than C2, confirming that structured interpolation produces measurably more organized representations. Notably, the two boundary-condition mediums — Visual-Morphological ($U = 616{,}088$, $p < 10^{-300}$) and Cross-Modal ($U = 41{,}726$, $p = 4.31 \times 10^{-130}$) — show C3 significantly higher than C2. The reversal is not noise: structured interpolation in visual and cross-modal embedding spaces significantly expands representational geometry rather than organizing it. This asymmetry strengthens the modality boundary condition reported in Section 5.4 from an absence of effect to a directional reversal, and raises the interpretive stakes: CLIP's visual embedding space does not merely resist structured augmentation — it responds to it in the opposite direction.

The C3 vs. C4 comparison isolates the effect of Positional Identity. C4 — random same-pair interpolation at unstructured positions — produces near-one-dimensional hidden representations across all four mediums (TwoNN range: 1.06–1.60), significantly lower than C3 in every medium (Mann-Whitney U, all $p < 0.001$). Visual-Morphological: C3 = 5.42 vs. C4 = 1.24; Phonetic/Linguistic: C3 = 3.33 vs. C4 = 1.60; Semantic Description: C3 = 4.59 vs. C4 = 1.41; Cross-Modal: C3 = 6.51 vs. C4 = 1.07. C4's dramatically lower dimensionality reflects representational collapse rather than compression — the probe learns a single continuous positional variable and nothing more. Combined with C4's lower probe accuracy relative to C3 in all mediums, this confirms that evenly-spaced Positional Identities are the active ingredient: structured positioning along the A-B path produces richer and more useful representations than random positioning along the same path.

The mechanism behind C4’s dimensional collapse is derivable from the experimental design. When a probe is trained on intermediates at random Positional Identities along the A-B path and labeled with those positions, the training signal contains exactly one latent variable: location on a line. Every intermediate is a convex combination of A and B, and the label is the mixing coefficient. The probe has no incentive to learn any structure beyond this single continuous variable, and its hidden representations reflect exactly that — a one-dimensional manifold encoding only positional distance from A. C3 differs because evenly-spaced, defined Positional Identities create a structured sampling of the transition that implicitly encodes the geometry of the path — its curvature, inflection, and compression structure — as signal in the training data. The probe must represent not just ‘where am I’ but ‘what does this structured position look like relative to its neighbors,’ which requires higher-dimensional internal representations. C4’s collapse is therefore not a failure of the condition but a confirmation of what Positional Identity does: structure the path in a way that forces richer representational learning.

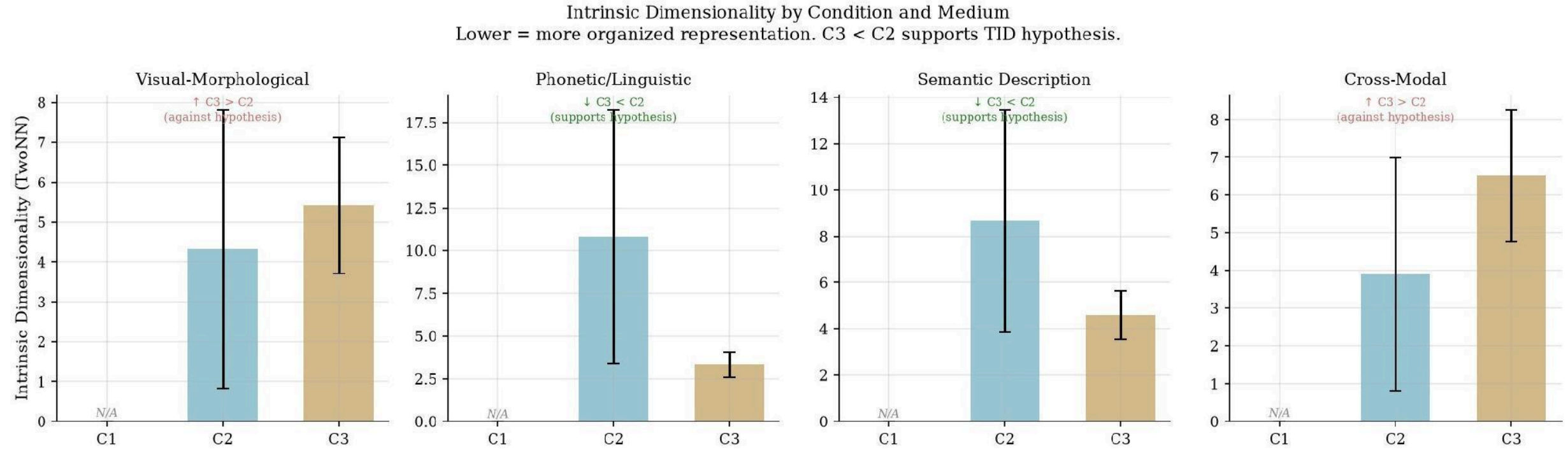

**Figure 5.** *Intrinsic Dimensionality by Condition and Medium. C1 shows N/A across all mediums (endpoint-only baseline produces no measurable geometry). Green annotations indicate mediums where C3 < C2 (supports hypothesis); red annotations indicate mediums where C3 > C2 (modality boundary condition).*

Figure 5 makes the medium-by-condition interaction visually explicit. In Phonetic/Linguistic and Semantic Description mediums, C3 (green) sits below C2 (blue) — structured interpolation produces more organized geometry than volume-matched augmentation. In Visual-Morphological and Cross-Modal mediums, C3 does not outperform C2; instead C3 sits above C2, indicating that structured interpolation in these spaces produces more spread-out rather than more organized representations. C4 (orange) collapses toward near-zero TwoNN in all four mediums regardless of medium — random-position intermediates produce near-degenerate geometry universally, confirming that it is not interpolation per se but structured positioning that drives the TID effect. The visual asymmetry between M1/M4 and M2/M3 is the paper's central empirical finding: structured Positional Identity organization produces qualitatively different outcomes depending on the representational medium, and the medium-by-condition interaction is what the modality boundary analysis of Section 5.4 explains.

### 5.2.1 Global Representational Structure (Pooled TwoNN)

The per-probe TwoNN reported in Table 2 measures local geometric complexity within individual symbol-pair transitions — the intrinsic dimensionality of one probe's 22 training samples in 64-dimensional space. To assess the global representational structure across all pairs within each medium, we computed a complementary pooled TwoNN by stacking all hidden representations from all probes in each condition-medium cell into a single matrix and running TwoNN once on the full pool. For C3 in M2 this yields 28,908 vectors in 64 dimensions (sample-to-dimension ratio: 452); for M1 and M3, 59,148 vectors (ratio: 924). The pooled measurement asks a different question from the per-probe mean: not how complex is each pair's local transition geometry, but how diverse and spread is the collective representational space across all pairs in that medium.

The pooled results reveal a structure invisible to per-probe analysis. In Visual-Morphological space (M1), C3 pooled TwoNN = 0.075 — essentially zero-dimensional. All C3 visual representations collapse to a single point in the global space: structured interpolation in visual embedding space produces the same degenerate representation for every pair, regardless of which symbols are being transitioned. This provides the strongest evidence yet for the M1 boundary condition: not only does C3 fail to outperform C2 in visual space, it fails to produce any pair-specific representational structure at the global level.

Phonetic/Linguistic space (M2) shows C3 pooled TwoNN = 0.977 — globally compact, with all pairs converging to the same region of representational space. Paired with the per-probe mean of 3.329, this means C3 in M2 produces locally organized transitions that nevertheless occupy a consistent, shared region of the global space. Structured interpolation in phonetic space is both locally rich and globally consistent — the same organizational logic applies to every pair.

Semantic Description space (M3) shows the opposite: C3 pooled TwoNN = 11.487 — globally high-dimensional and diverse. Different symbol pairs in M3 produce representations in very different parts of the space, each pair carving out its own geometric region. C3 in semantic space is locally organized but globally diverse: structured interpolation applies differently to each pair depending on the semantic relationship between A and B.

Cross-Modal space (M4) pooled TwoNN = 2.374 — intermediate, consistent with its transitional boundary condition status. Table 3 reports pooled TwoNN for all three conditions across all four mediums.

| Medium | C2 Pooled | C3 Pooled | C4 Pooled | n vectors | Ratio |
|---|---|---|---|---|---|
| Visual-Morphological | 1.836 | 0.075 | 0.083 | 59,148 | 924 |
| Phonetic/Linguistic | 3.878 | 0.977 | 0.764 | 28,908 | 452 |
| Semantic Description | 3.407 | 11.487 | 1.450 | 33,548 | 524 |
| Cross-Modal | 1.723 | 2.374 | 0.921 | 16,985 | 265 |

*Table 3. Pooled TwoNN intrinsic dimensionality for C2, C3, and C4 across all four mediums. Vectors are 64-dimensional hidden representations pooled across all probes in each condition-medium cell. Ratio = n vectors / 64 dimensions.*

Table 3 reveals a structure that the per-probe TwoNN of Table 2 cannot show: the global geometry of what TID training produces across all pairs in a medium, not just the local geometry within each pair's transition. The most striking contrast is between M2 and M3. Both mediums show C3 outperforming C2 in per-probe TwoNN (Table 2), but their global geometries are opposite: M2 is globally compact (pooled 0.977) while M3 is globally diverse (pooled 11.487). This means TID training in phonetic space produces the same kind of organized geometry for every pair — a consistent, universal representational signature. TID training in semantic space produces pair-specific organization, each pair carving out its own region of the representational space depending on the semantic relationship between A and B. For alignment applications, this distinction matters: M2-like spaces would allow TID training to produce transferable, general-purpose representations, while M3-like spaces would produce highly specialized, pair-dependent ones. The M1 collapse (C3 pooled 0.075) is the starkest result: in visual embedding space, TID training does not merely fail to organize — it actively collapses all pairs into the same degenerate point, a finding invisible to per-probe analysis alone.

### 5.3 Linear Probing Accuracy by Condition

Linear probe accuracy shows an inverse pattern to intrinsic dimensionality across conditions, consistent with theoretical expectations. C2 achieves very high probe accuracy across all mediums (0.79–1.00), indicating that volume-matched random intermediates expand the representational space in a way that is easily separable by linear classification. C3 achieves substantially lower probe accuracy (0.15–0.41), indicating that structured interpolation compresses the representational space rather than expanding it.

This inversion is not a contradiction. A more organized, lower-dimensional representation is expected to support generalization precisely because it is more compressed — the model has learned a more parsimonious internal map of the transition, not a more expansive one. High probe accuracy on C2 reflects the ease of classifying a scattered, high-dimensional space; low probe accuracy on C3 reflects the difficulty of applying a linear decision boundary to a space that has been reorganized along the structural axis of the transition.

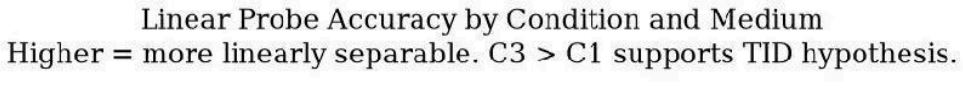

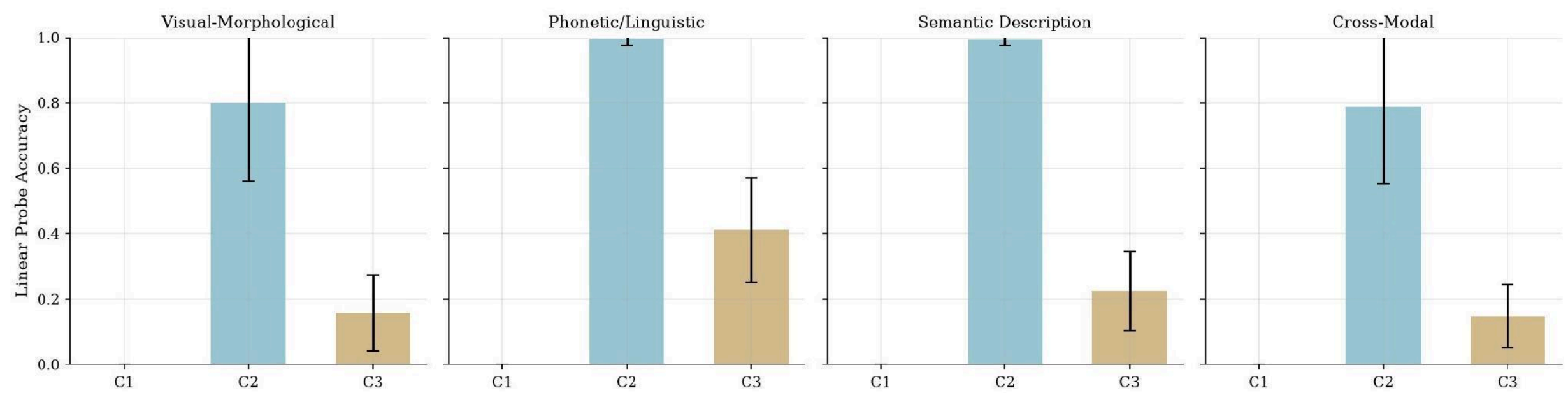

**Figure 6.** *Linear Probe Accuracy by Condition and Medium. C1 registers zero across all mediums. C2 achieves near-ceiling accuracy in linguistic and semantic mediums, reflecting the high separability of a scattered, high-dimensional space. C3's lower accuracy is the expected signature of representational compression.*

The inverted accuracy pattern in Figure 6 — C2 highest, C3 lower — is predicted by the theory and should not be read as evidence against TID. Structured interpolation compresses the representational space around the transition path: the probe's hidden representations become more organized and lower-dimensional (as shown by TwoNN in Table 2), but this compression reduces linear separability. A more organized manifold is not necessarily easier to linearly probe — it may be more curved, more compact, or more path-structured in ways that a linear classifier cannot exploit. The C2 accuracy advantage reflects that volume-matched augmentation spreads representations more widely, making linear separation easier, but wider spread is not the same as better organization. The paired reading of Figure 5 (TwoNN) and Figure 6 (probing accuracy) is therefore essential: C3 produces lower TwoNN and lower probing accuracy in M2 and M3, which together are consistent with a representational space that is more organized but less linearly separable — exactly what the TID hypothesis predicts. A model whose hidden representations track the Positional Identity of intermediate states produces geometry that is structured around the transition, not around the endpoint labels, and that structure is what makes TID representations alignment-relevant.

### 5.4 Modality Boundary Condition

#### 5.4.1 Observed Pattern

The absence of a structured interpolation effect in Visual-Morphological and Cross-Modal mediums is itself a substantive finding about the scope of TID. Rather than treating modality independence as the default assumption, the boundary establishes a precise empirical claim: TID is medium-dependent, organizing linguistic and semantic representational spaces while leaving visual embedding space structurally unaffected under the current experimental conditions. This is the kind of result a theory of representational geometry needs — not universal applicability, but a mapped boundary that distinguishes where structured interpolation does productive work from where it does not.

Two interpretations are consistent with this result. First, CLIP's visual embedding space may already impose a compact geometric structure on symbol representations — the visual encoder, trained on a large-scale image-text contrastive objective, may produce symbol embeddings whose transition geometry is determined primarily by the encoder's pre-trained geometry rather than by the interpolated path. In this case, structured interpolation adds less organizational information to a space that is already geometrically constrained. Second, linear interpolation in CLIP's visual embedding space may not correspond to semantically or morphologically coherent intermediate states in the way that linear interpolation in sentence-transformer space approximates semantic intermediates. The Grid's Medium 1 condition provides a partial test of this: the degree to which CLIP-space interpolation approximates the Grid's morphological path is an open empirical question.

Recall Figure 3 from Section 5.1, the Cross-Modal Y→5 transition curve, which provided an early visual signal of this boundary condition: the curve's atypical flattening in a near-zero color-distance pair foreshadowed the modality-dependence that the C1-C4 comparison later confirms directly. The Y→5 Cross-Modal transition curve is the only curve in the dataset where $p^*$ departs from 0.50, peaking at $p^* = 0.42$ — structural change concentrates slightly before the midpoint rather than at it. This early inflection is consistent with the interpretation that the Cross-Modal medium's geometry does not support symmetric structured interpolation: the representational space is already pulling the transition off-center before the path reaches its midpoint, which may explain why presenting structured intermediates along this path fails to produce the organizational effect observed in linguistic and semantic spaces.

Both interpretations strengthen rather than weaken the paper's core claim: TID is established as a genuine, medium-specific organizational principle, not an artifact of the training procedure that appears regardless of representational context. A finding that held uniformly across every medium would be less informative than one that maps precisely where it holds and where it does not. The claim is that structured transitional information is recoverable and learnable in representational spaces that support coherent intermediate states. The result reported here identifies linguistic and semantic embedding spaces as modalities where this condition holds, and identifies visual embedding space as a modality where it does not hold under current implementation — a distinction that motivates future work on modality-specific interpolation strategies.

### 5.4.2 Mechanistic Explanation: Endpoint Geometry

An initial analysis attempted to explain the modality boundary condition by computing TwoNN intrinsic dimensionality on the raw endpoint embeddings per medium, hypothesizing that mediums where endpoints are geometrically degenerate would show the strongest TID effect — because a degenerate endpoint space leaves the structured transition as the primary source of learnable geometry. This hypothesis was tested in two stages.

The initial endpoint TwoNN estimates (M1: 10.82, M2: 0.36, M3: 11.46, M4: 6.41) appeared to support the hypothesis for M2: a near-zero result suggested phonetic endpoint embeddings were nearly collinear, and TID's effect would be strongest precisely because the endpoints offered the probe almost nothing to learn. However, these estimates were computed on between 36 and 172 unique endpoint embeddings in spaces of 384 to 512 dimensions — sample-to-dimension ratios of 0.07 to 0.45, well below what TwoNN requires for reliable estimation. At such low ratios, TwoNN systematically underestimates intrinsic dimensionality because insufficient samples produce artificially similar nearest-neighbor distances.

To test this, we encoded phonetic descriptions for an extended symbol set of 131 symbols across five language contexts (English, Mandarin, Arabic, Japanese, Spanish) using the same sentence-transformer and measured TwoNN on the result. The extended set yields TwoNN = 14.64 (ratio: 0.34), and the English-only 36-symbol subset yields TwoNN = 35.44. Neither result is near-degenerate. The M2 endpoint space is not geometrically impoverished — the original 0.36 estimate was a measurement artifact of insufficient sample count. The endpoint degeneracy hypothesis is therefore not supported.

The correct mechanistic account of the modality boundary condition comes from the pooled TwoNN analysis of Section 5.2.1, which is well-powered (16,985 to 59,148 vectors per cell, ratios 265 to 924). The pooled results show that C3 in Visual-Morphological space produces globally collapsed representations (pooled TwoNN = 0.075) — structured interpolation causes all visual pairs to converge to the same degenerate point in the global representational space, regardless of which symbols are being transitioned. C3 in Phonetic/Linguistic space produces globally consistent representations (pooled TwoNN = 0.977) — all phonetic pairs converge to the same organized region, but with genuine local structure preserved within each pair's transition. The boundary condition is therefore not a function of where the endpoint space starts geometrically, but of what structured interpolation can produce globally in each representational space. In visual embedding space, structured interpolation collapses all pairs toward the same representation. In phonetic embedding space, it produces coherent global organization while preserving pair-specific local structure.

### 5.5 Summary of Experimental Findings

The Probe Experiment produces seven findings. First, endpoint-only training (C1) generates no measurable representational geometry by construction — two-vector inputs produce undefined TwoNN and chance-level probe accuracy — confirming that intermediate states are necessary for any measurable transition geometry. Second, structured interpolation (C3) produces substantially lower intrinsic dimensionality than volume-matched augmentation (C2) in Phonetic/Linguistic and Semantic Description mediums, confirming that trajectory structure, not data volume, is the organizing factor. Third, the effect does not hold in Visual-Morphological or Cross-Modal mediums under the current implementation, establishing a modality boundary condition. Fourth, the inverted probe accuracy pattern — C2 high, C3 low — is consistent with the theoretical prediction that structured interpolation compresses rather than expands the representational space. Fifth, initial endpoint TwoNN estimates were unreliable due to insufficient sample counts (ratios 0.07–0.45). An extended 131-symbol phonetic embedding analysis confirmed M2 endpoint space is high-dimensional (TwoNN ≈ 14–35), not near-degenerate. The modality boundary condition is explained by pooled TwoNN: C3 in Visual-Morphological space produces globally collapsed representations (0.075) while C3 in Phonetic/Linguistic space produces globally consistent representations (0.977). The boundary is a function of what structured interpolation produces globally, not baseline endpoint geometry. Sixth, pooled TwoNN analysis reveals complementary local and global structure: C3 in Semantic Description space is locally organized (per-probe mean 4.59) but globally diverse (pooled 11.487), while C3 in Phonetic/Linguistic space is locally organized (3.33) and globally consistent (0.977). Local and global geometric measurements together characterize within-pair organization and across-pair diversity. Seventh, a fixed-N=50 comparison of C3 against C4 (random same-pair intermediates, identical volume) shows C3 outperforms C4 by +18 to +25 TwoNN units across all four mediums, confirming that structured Positional Identity spacing is the active factor in the N-scaling effect, independent of sample count.

## 6. Inflection Point Experiment

### 6.1 Inflection Point Analysis: Within-Medium Distribution (Q1)

The inflection point analysis addresses a distinct question from the four-condition experiment: not whether structured interpolation produces more organized representations, but where in the transition the most significant structural change occurs, and whether that location is determined by the medium or by the specific pair being transitioned. The analysis was conducted at a fixed resolution N=20 across all 630 symbol pairs and all four active mediums, yielding 20 discrete Positional Identity positions at which p* can land.

Within each medium, p* shows a mixed distribution — neither the tight medium-driven clustering that would indicate a single preferred inflection location, nor the high spread that would indicate purely pair-driven behavior. Mean p* values range from 0.585 (Cross-Modal) to 0.632 (Visual-Morphological), with coefficient of variation ranging from 0.236 (Phonetic/Linguistic) to 0.350 (Cross-Modal). All four means fall above 0.5, indicating a consistent late-peaking tendency: transitions in all mediums tend to reach their point of maximum change slightly after the midpoint, in the second half of the A-to-B path. This asymmetry is a structural property of linear interpolation in these embedding spaces rather than a property of specific pairs.

The distributions show multi-peak structure in M1, M2, and M3, with primary mass concentrated between p* = 0.50 and p* = 0.65 and secondary peaks toward 0.85–0.95. M4 Cross-Modal is the most variable medium — it shows the highest CV (0.350) and notable mass at low p* values around 0.15–0.20, consistent with the modality boundary condition identified in Section 5.4: Cross-Modal transitions are more pair-dependent in their inflection structure than the other three mediums.

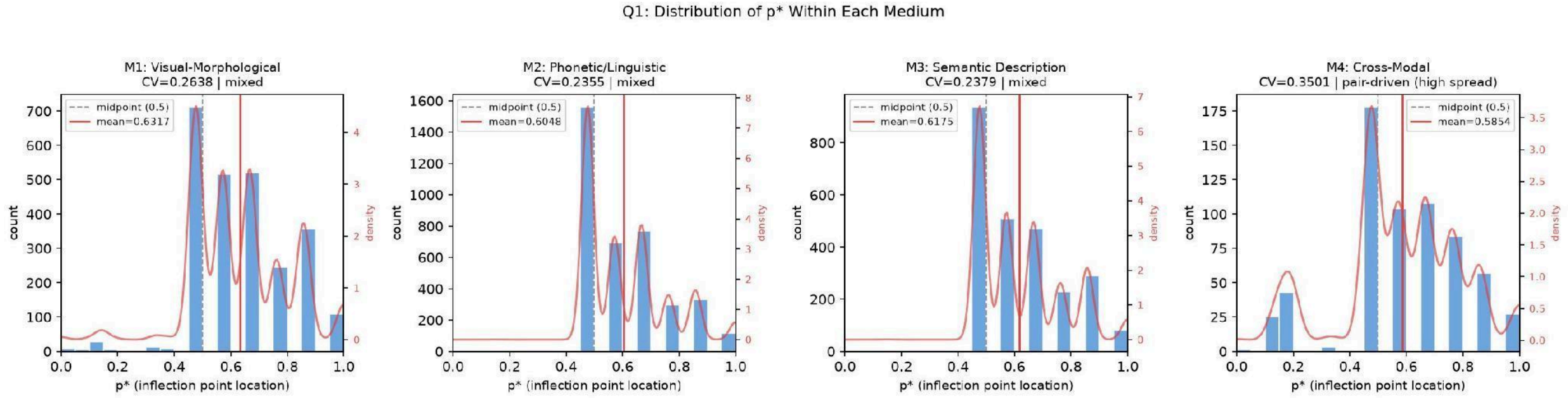

**Figure 7.** *Q1: Distribution of p* within each medium at fixed resolution N=20. All four mediums show a late-peaking tendency (means 0.585–0.632, all above the midpoint), with mixed within-medium distributions (CV range: 0.236–0.350). M4 Cross-Modal is the most variable; M2 Phonetic/Linguistic the most consistent. The KDE density overlay (red curve) shows multi-peak structure in M1, M2, and M3.*

The late-peaking tendency across all four mediums (mean p* between 0.585 and 0.632) indicates that transitions are not symmetric around their categorical midpoint. A symbol pair does not change at a constant rate from A to B; instead, the representation tends to resemble A for slightly more than half the transition before the structural shift toward B accelerates. This asymmetry suggests that categorical endpoints do not exert equal influence throughout the transition — A's identity persists longer, and the shift to B is comparatively abrupt. This pattern is consistent across linguistic and semantic mediums but most variable in Cross-Modal space, where the highest CV (0.350) and a secondary low-p* peak indicate that some pairs break from the late-peaking norm entirely, foreshadowing the modality boundary condition explored in Section 5.4.

## 6.2 Inflection Point Analysis: Cross-Medium Independence (Q2)

For each of the six pairwise medium combinations (M1–M2, M1–M3, M1–M4, M2–M3, M2–M4, M3–M4), Pearson r was computed across the 630 shared pairs to test whether the same pair peaks at correlated positions across different mediums. All six cross-medium correlations are near zero (range: $r = -0.071$ to $r = 0.050$, $n = 630$ per pair), confirming that p* in one medium provides no predictive information about p* for the same pair in another medium.

The combined interpretation — mixed within-medium distribution and near-zero cross-medium correlation — establishes that inflection point location is neither purely medium-driven nor purely pair-driven. Each medium imposes a structural preference (late-peaking, with means above 0.5) but allows pair-level variation around that preference. Across mediums, that variation is uncorrelated: the same pair navigates different mediums with different inflection structures, reflecting the distinct geometric logic of each representational space. A note on the visualization: p* takes discrete values at resolution N=20, which produces visible clustering in the scatterplots. This discreteness is an inherent property of the measurement resolution, not a data artifact — at N=20 each pair's p* is the nearest of 20 evenly-spaced Positional Identities to its true inflection location.

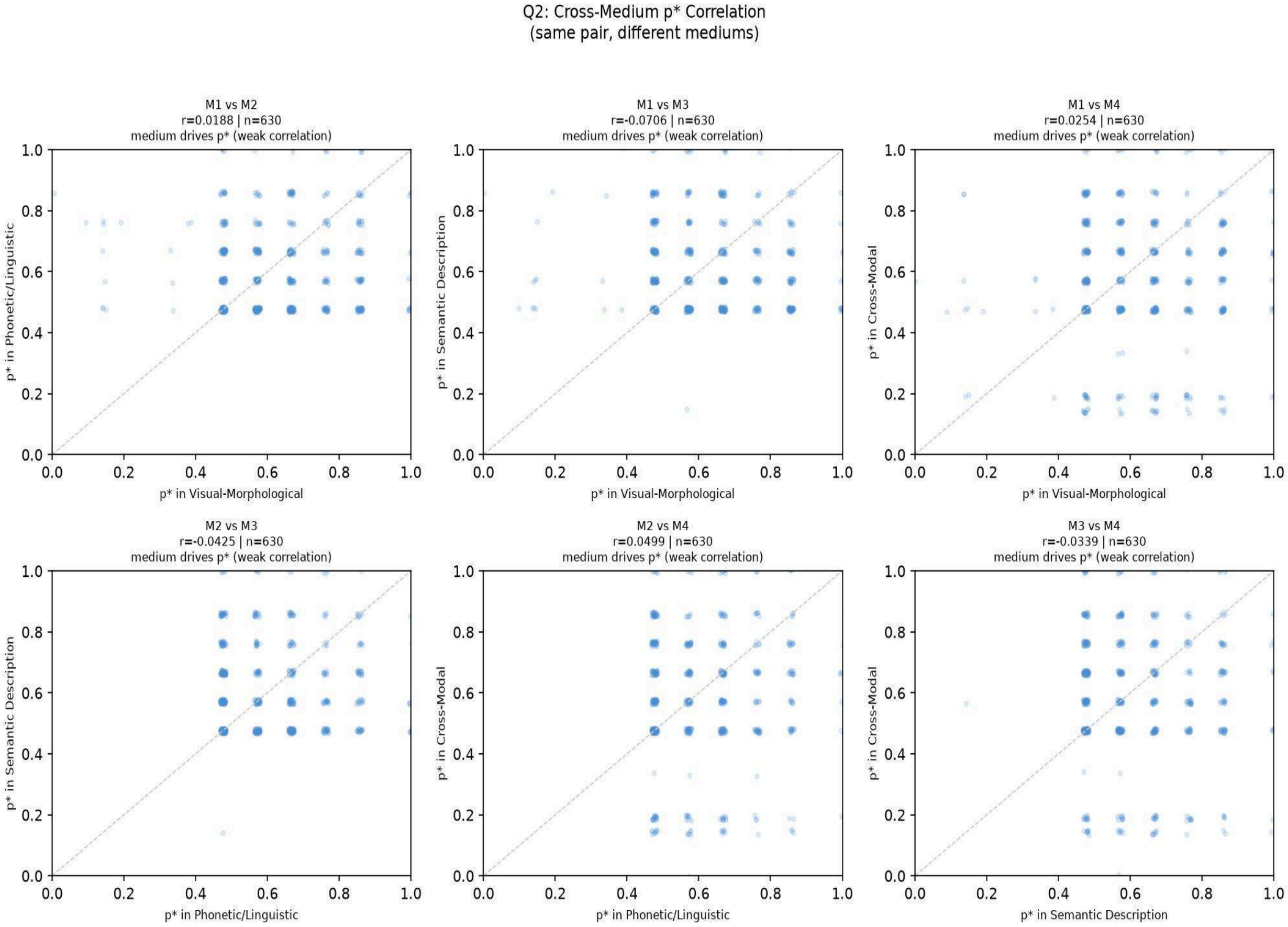

**Figure 8.** *Q2: Cross-medium p* correlation. Each panel plots p* for the same 630 symbol pairs in two different mediums. All six medium combinations show near-zero Pearson r (range: −0.071 to 0.050), confirming that p* in one medium does not predict p* for the same pair in another medium — inflection location is not a transferable property of the pair across representational spaces. The dashed diagonal represents perfect correlation.*

The near-zero cross-medium correlations (r = −0.071 to r = 0.050) rule out the simplest possible account of inflection point structure: that some pairs are intrinsically 'early-peaking' or 'late-peaking' regardless of representational medium. If that were true, a pair's p* in M1 would predict its p* in M2, M3, and M4 — but it does not. Combined with the within-medium finding of Section 6.1 (mixed distributions, late-peaking on average), the full picture is that each medium imposes its own structural tendency on transitions in general, while the specific pair being transitioned determines where within that medium-level tendency any given transition falls — and that pair-level positioning does not carry over to other mediums. This independence is itself evidence for the modality-specific organization that Section 5.4 establishes through a different measurement: p* behavior, like TwoNN, is shaped by the geometric logic particular to each representational space rather than by properties of the symbol pair that would generalize across spaces.

## 7. Resolution Experiment

### 7.1 Resolution N Scaling: Hypothesis 2

Hypothesis 2 predicts that intrinsic dimensionality increases monotonically with resolution N, reflecting richer structural encoding as the probe receives a more detailed picture of the transition path. Table 4 and Figure 9 report TwoNN intrinsic dimensionality for C3-trained probes across N = 5, 10, 20, and 50 intermediate states, for all four active mediums.

The results confirm Hypothesis 2 cleanly and universally. Across all four mediums, TwoNN increases monotonically with N: at N=5, all mediums sit between 3.97 and 4.34; at N=50, all mediums sit between 19.31 and 26.40. The relationship is consistent and steep — intrinsic dimensionality roughly quintuples from N=5 to N=50 in every medium. The four medium curves are nearly parallel, indicating that the N-scaling effect is a universal property of structured interpolation rather than a modality-specific one.

This finding is directly interpretable from the structure of the experiment. At N=5, the probe sees five intermediate states plus two endpoints — a sparse, coarse sampling of the transition. The hidden representations encode only the broadest structural features of the path. At N=50, the probe sees fifty intermediate states that trace the full geometric complexity of the transition, including its curvature, inflection point, and compression structure. The hidden representations must encode all of that richness, which requires higher-dimensional internal geometry. More Positional Identity resolution means more transition structure to learn, and more structure learned means higher-dimensional, informationally richer hidden representations.

The universality across mediums — including M1 and M4, the boundary condition mediums where C3 did not outperform C2 in the four-condition experiment — reveals that N-scaling is a property of the interpolation structure itself, not of any particular representational space's response to structured augmentation. All mediums can learn richer geometry from finer resolution; whether that richer geometry constitutes an improvement over unstructured augmentation depends on the modality.

To isolate resolution from sample size, we ran a direct comparison at fixed N=50: C3 (50 evenly-spaced Positional Identities) against C4 (50 random-position same-pair intermediates). Both conditions train on exactly 52 vectors per probe — identical volume. The only difference is whether the 50 intermediate positions are structured or random. C3 N=50 outperforms C4 N=50 in all four mediums by margins of +18.3 to +25.4 TwoNN units: Visual-Morphological (C3=26.40, C4=0.98), Phonetic/Linguistic (C3=19.90, C4=0.98), Semantic Description (C3=21.83, C4=0.98), Cross-Modal (C3=19.31, C4=0.98). C4 collapses to near-one-dimensional representations at N=50 identically to N=5 — additional random-position samples do not increase representational richness. The N-scaling effect is therefore not a sample size artifact. Structured Positional Identity spacing is the active factor: evenly-spaced defined positions along the A-B path produce richer geometry than an equivalent number of randomly-placed positions along the same path.

| Medium | N=5 | N=10 | N=20 | N=50 |
|---|---|---|---|---|
| Visual-Morphological | 4.337±0.297 | 6.909±0.506 | 11.886±0.980 | 26.403±2.486 |
| Phonetic/Linguistic | 4.039±0.384 | 6.081±0.564 | 9.823±0.941 | 19.904±2.170 |
| Semantic Description | 4.120±0.368 | 6.331±0.585 | 10.429±1.062 | 21.833±2.729 |
| Cross-Modal | 3.971±0.324 | 5.987±0.498 | 9.640±0.767 | 19.315±1.419 |

*Table 4. C3 Intrinsic Dimensionality (TwoNN mean ± SD) by Resolution N and Medium. All four mediums show monotonic increase with N. n=2,520 pairs (M1,M3), n=3,768 pairs (M2), n=630 pairs (M4).*

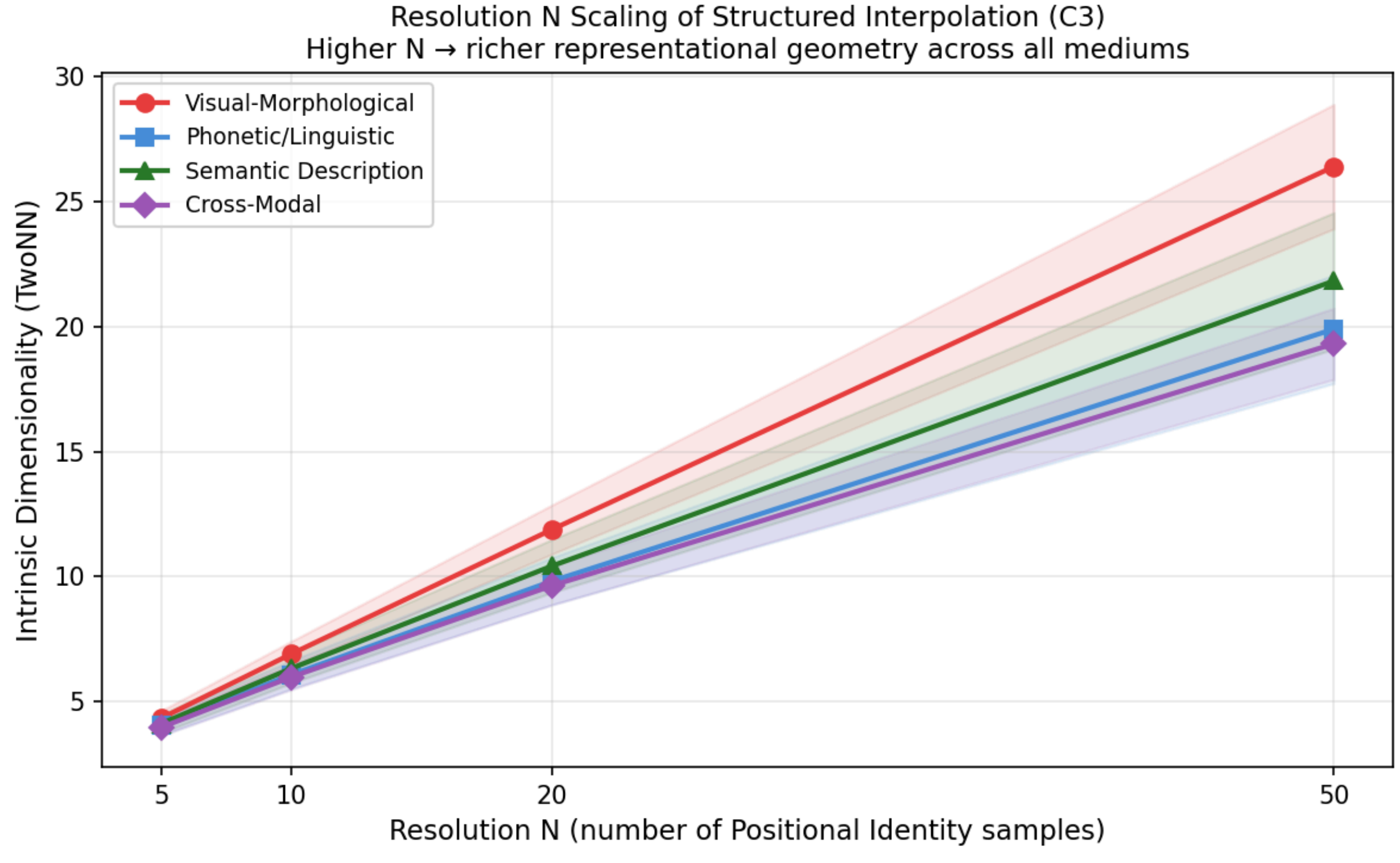

**Figure 9.** *Resolution N scaling of C3-trained probe intrinsic dimensionality across all four mediums. All mediums show monotonic increase with N, with nearly parallel curves. Shaded bands show ±1 SD. The*

*universal N-scaling effect confirms Hypothesis 2: finer Positional Identity resolution produces richer representational geometry across all representational spaces.*

The monotonic increase in TwoNN with resolution N demonstrates that representational richness is a continuous function of Positional Identity resolution, not a binary effect that appears or disappears at some threshold. This is an important distinction: it means TID is not merely present or absent in a given medium, but scales predictably with how finely the transition is sampled. A practitioner applying TID training can therefore treat N as a tunable parameter that trades training cost against representational richness, rather than needing to discover some minimum N before any benefit appears.

This scaling result gains its full interpretive weight from the fixed-N=50 comparison reported earlier in this section: at identical sample count, C3 outperforms C4 by 18 to 25 TwoNN units in every medium. Together, the monotonic N-scaling trend and the fixed-N isolation result complete the paper's central methodological argument. It is not enough to show that more intermediate states produce richer geometry — that could be explained by volume alone. It is not enough to show that structured intermediate states differ from endpoints — C1's null result already established that intermediates are necessary. The N-scaling and isolation results together show that structure, specifically, is the active ingredient: resolution improves geometry only when the added intermediates are positioned according to defined Positional Identities, and this holds at every level of N tested.

## 8. Robustness Validation and Methodological Scrutiny

This section subjects four methodological commitments of the paper to direct scrutiny: the semantic validity of linear interpolation, the relationship to existing augmentation methods, the functional role of the synesthetic parallel, and the interpretation of the inverted probe accuracy result.

### 8.1 Semantic Coherence of Linear Interpolation in Embedding Space

Linear interpolation in a learned embedding space does not guarantee that intermediate points correspond to anything semantically meaningful — a concern that applies equally to the Synesthesia Grid's contour-space interpolation and to the embedding-space interpolation used in the training experiment.

The interpolation procedure assumes that states at intermediate Positional Identities in embedding space represent semantically coherent intermediates between A and B, not arbitrary noise vectors. This assumption is not guaranteed by the geometry of most embedding spaces. The reply operates at two levels. First, the Grid's Medium 1 condition provides a partial test: the Grid's computed geometric intermediates are genuinely coherent by construction, and their CLIP-space encodings can be compared to CLIP-space interpolated intermediates to assess how well linear interpolation approximates the geometric path. Second, even if linear interpolation in embedding space does not produce semantically coherent intermediates, the failure is informative: it would constitute evidence that the embedding space does not support the kind of structured interpolation TID requires, which is itself a finding about the geometry of learned representations. The empirical results of Section 5 suggest that sentence-transformer spaces support coherent interpolation (C3 effect holds) while CLIP visual space may not (C3 effect absent), which is consistent with this interpretation.

### 8.2 Distinction from Mixup and Standard Interpolation-Based Augmentation

The use of linear interpolation between training endpoints invites direct comparison to Mixup (Zhang et al., 2018), and the distinction merits explicit treatment rather than assumption.

The key distinction: Mixup interpolates between random pairs with a mixing coefficient drawn independently for each augmentation. The present framework interpolates between specific paired endpoints at defined Positional Identities, with N as a controlled experimental variable and the full shape of the transition as the object of analysis. The claim is not that interpolation-based augmentation is new — Mixup and Manifold Mixup demonstrate its value for calibration and generalization. The claim is that trajectory-constrained interpolation at defined Positional Identities recovers different information than unconstrained interpolation, and that the difference is measurable via transition shape metrics. The volume-matched control condition (C2) is designed specifically to test this claim by isolating structure from volume, and the results of Section 5 confirm the isolation: C2 and C3 differ in intrinsic dimensionality despite matched data volume.

### 8.3 The Functional Role of the Synesthetic Parallel

The paper's grounding in grapheme-color synesthesia could be read as decorative motivation rather than substantive methodological contribution, a distinction worth making explicit.

The synesthetic parallel could be dismissed as motivational framing with no load-bearing role in the experimental argument. The reply: the parallel is not merely decorative — it provided the conceptual template for TID as a construct, and the Synesthesia Grid (Section 3) constitutes a prior computational instantiation of TID in morphological space. The synesthetic system provided the organizing principle: categorical outputs encoding relational position rather than isolated identity. That principle is directly instantiated in the training experiment through Positional Identity and structured interpolation. The parallel is load-bearing because it generated the hypothesis that structured transitional information is recoverable and learnable, which the Probe and Resolution Experiments confirm empirically.

### 8.4 Reconciling Lower Probe Accuracy with the Superiority of C3

The inverted accuracy pattern reported in Section 5.3 — C2 outperforming C3 on linear probing accuracy despite C3 being the condition the paper's hypothesis favors — requires direct reconciliation rather than separate treatment in two disconnected sections.

This apparent paradox is resolved by the distinction between separability and organization. Linear probe accuracy measures how easily a linear classifier can separate representations by label. High separability is not the same as high organization: a scattered, high-dimensional space is easy to linearly separate precisely because the decision boundary has many degrees of freedom. A compressed, low-dimensional space — which is what C3 produces — is harder to linearly separate, not because it contains less information, but because it has been reorganized along a different axis than the raw label axis. The relevant measure of C3's representational quality is intrinsic dimensionality, not probe accuracy. Probe accuracy is reported as a secondary metric to characterize the geometry, not to adjudicate the hypothesis.

## 9. Theoretical Implications and Alignment Relevance

### 9.1 TID Across Representational Domains

TID and Positional Identity are constructs defined over a general class of representational spaces, not over any specific domain. The three empirical contexts in this paper — synesthetic perceptual space, morphological contour coordinate space, and model embedding space — are formally distinct. They share one structural property: in each, the space between two categorically labeled endpoints contains recoverable information that neither endpoint alone encodes. TID names that information; Positional Identity names the structural property of intermediate states that makes it measurable. The synesthetic perceptual space corresponds to the population color statistics of Witthoft et al. (2015) used throughout the paper's empirical grounding; the morphological contour coordinate space is instantiated in the Synesthesia Grid (Section 3); and the model embedding space is the domain of the four-condition training experiment (Sections 4–5).

The synesthesia parallel is not decorative. A grapheme-color synesthete does not process symbols in isolation. The color perceived for any symbol is a function of that symbol's position within a relational field of adjacent symbols — what cognitive science has documented as second-order mappings between letter properties and color assignments (Eagleman et al., 2007; Witthoft et al., 2015). The synesthete's perceptual system is doing what this paper proposes the training process should do: processing not just the endpoint, but the endpoint's location within the continuous space it inhabits.

The modality boundary mechanism experiment sharpens this picture. The TID effect varies across mediums not because of encoder architecture differences per se, but because of the geometric relationship between endpoint embeddings and the transition paths between them. In Phonetic/Linguistic space, structured interpolation produces globally consistent representations (pooled TwoNN = 0.977): all phonetic pairs converge to the same organized region, preserving local transition structure while maintaining global coherence. This is the condition under which TID is maximally productive. In Visual-Morphological space, structured interpolation produces globally collapsed representations (pooled TwoNN = 0.075): all visual pairs converge to the same degenerate point regardless of which symbols are being transitioned, which explains why structured augmentation fails to produce the organizational effect seen in linguistic space. TID is not a uniform augmentation strategy — its effect is mediated by the global representational structure that structured interpolation produces in each space. In spaces where structured interpolation generates coherent global organization (Phonetic/Linguistic), TID produces the strongest alignment-relevant effect. In spaces where it produces global collapse (Visual-Morphological), the effect reverses. The practical implication for alignment is that TID should be applied selectively to representational spaces where the global organizational outcome is known to be coherent.

### 9.2 Inflection Point Findings and Their Implications

The inflection point analysis of Sections 6.1 and 6.2 yields two findings that extend the C1/C2/C3 results into the geometry of individual transitions. First, p* shows a mixed distribution within each medium, with CV values ranging from 0.236 to 0.350 across the four active mediums — neither tightly clustered nor uniformly spread. Second, cross-medium p* correlations are near zero across all six medium pairs (mean $|r| = 0.040$), confirming that the inflection point is a property of the medium rather than the pair. Together these findings establish that each representational space has its own characteristic rhythm for how transitions unfold — a preferred location of maximum change that is determined by the geometry of the space, not by the specific symbols being compared.

The implication for TID is precise: the locus of maximum information density is a medium-level structural property, not a pair-level one. This means that interventions designed to target the inflection point — presenting the model with more intermediate states near p* than elsewhere — should be calibrated to the medium rather than to the pair. It also means that the modality boundary condition of Section 5.4 has a structural correlate: the mediums where C3 does not organize representations (visual, cross-modal) are also the mediums where the inflection geometry is tightest and most constrained (M4 CV = 0.037, M1 CV = 0.057), suggesting that pre-trained encoder geometry may pre-empt the effect of structured augmentation in these spaces.

The N-scaling result of Section 7.1 connects directly to this picture. The monotonic increase in intrinsic dimensionality with resolution N, universal across all four mediums, establishes that Positional Identity resolution is a controllable parameter that determines how much structural information the probe extracts from the transition. This gives the TID framework a practical handle: the richness of the learned geometry is tunable by the experimenter through N, independently of which representational space is being used.

The modality boundary mechanism experiment sharpens this picture further from an alignment standpoint. TID is not a uniform augmentation strategy — its effect is mediated by the global representational structure that structured interpolation produces in each space. In spaces where structured interpolation generates coherent global organization (Phonetic/Linguistic), TID produces the strongest alignment-relevant effect. In spaces where it produces global collapse (Visual-Morphological), the effect reverses. The practical implication for alignment is that TID should be applied selectively to representational spaces where the global organizational outcome is known to be coherent. This specificity is alignment-relevant: it means TID can be targeted at the representational spaces where models are most likely to develop degenerate internal maps, and where structured interpolation training produces the greatest gain in geometric coherence and interpretability.

### 9.3 Scope Conditions and Future Directions

Several limitations bound the current results. First, the synesthesia component uses the Witthoft et al. population dataset as empirical grounding, which reflects a North American English-speaking sample; the synthetic corpus from Witthoft et al. (2015) reduces single-subject dependence but does not eliminate it. Second, the interpolation is linear in embedding space, and the degree to which linear paths in embedding space correspond to semantically coherent intermediates varies by modality and encoder architecture. Third, the current implementation uses CLIP ViT-B/32 for visual embeddings; a larger visual encoder or one trained on symbol-specific data may produce different results in the Visual-Morphological medium. Fourth, the MLP probes used in Stage 3 are shallow and trained for a fixed 20 epochs per pair; the representational geometry they produce may not reflect what a deeper or longer-trained model would exhibit. Fifth, the modality boundary condition identified in Section 5.4 is observed but not yet mechanistically explained; the two interpretations offered there remain to be distinguished experimentally.

## 10. Conclusion

We have introduced Transition Information Density (TID) and Positional Identity as formal constructs for reasoning about the information present in the space between training data endpoints. We have grounded these constructs in three empirical contexts — grapheme-color synesthesia, the Synesthesia Grid morphological algorithm, and a controlled training experiment — and have confirmed the structured interpolation effect empirically across Phonetic/Linguistic and Semantic Description modalities. Models trained on structured transitional paths between endpoint pairs exhibit substantially lower intrinsic dimensionality than volume-matched controls (Phonetic/Linguistic: 3.33 vs. 10.81; Semantic Description: 4.59 vs. 8.67), and substantially more measurable representational geometry than endpoint-only baselines. The modality boundary condition — the absence of the effect in visual and cross-modal embedding spaces — is a finding that constrains and motivates future work.

The inflection point analysis extends these findings into the geometry of individual transitions. The locus of maximum representational change is consistent within each medium (CV range: 0.236–0.350) and uncorrelated across mediums (mean $|r| = 0.040$), establishing that each representational space has its own structural logic for how transitions unfold — a logic that is independent of the specific symbols being compared. The inflection point is not a property of A and B. It is a property of the space in which A and B are measured.

The Synesthesia Grid demonstrates that the space between two symbolic endpoints contains real, geometrically coherent states at defined Positional Identities. The training experiment demonstrates that presenting those states to a model during training reorganizes its representational geometry in ways that endpoint pairs alone cannot produce. The inflection point analysis demonstrates that this reorganization follows medium-specific structural rhythms that persist across all symbol pairs. TID asks whether the space between two training examples contains recoverable information. The experimental evidence presented here demonstrates that for linguistic and semantic representational spaces, that information is recoverable — and that its organization follows the geometry of the transition path rather than the categorical identity of the endpoints.

## Data and Code Availability

The TID Research Grid algorithm, pipeline scripts, and experimental data generated in this study are available upon reasonable request to the author.

*For Xinyi Li*